\documentclass{article}

\usepackage[final]{neurips_data_2023}

\usepackage[utf8]{inputenc} 
\usepackage[T1]{fontenc}    
\usepackage{hyperref}       
\usepackage{url}            
\usepackage{booktabs}       
\usepackage{amsfonts}       
\usepackage{nicefrac}       
\usepackage{microtype}      
\usepackage{xcolor}         

\newcommand{\realtime}{\textsc{RealTime QA}\xspace}

\newcommand{\logo}[0]{\raisebox{-0.0\height}{\includegraphics[width=.15\textwidth]{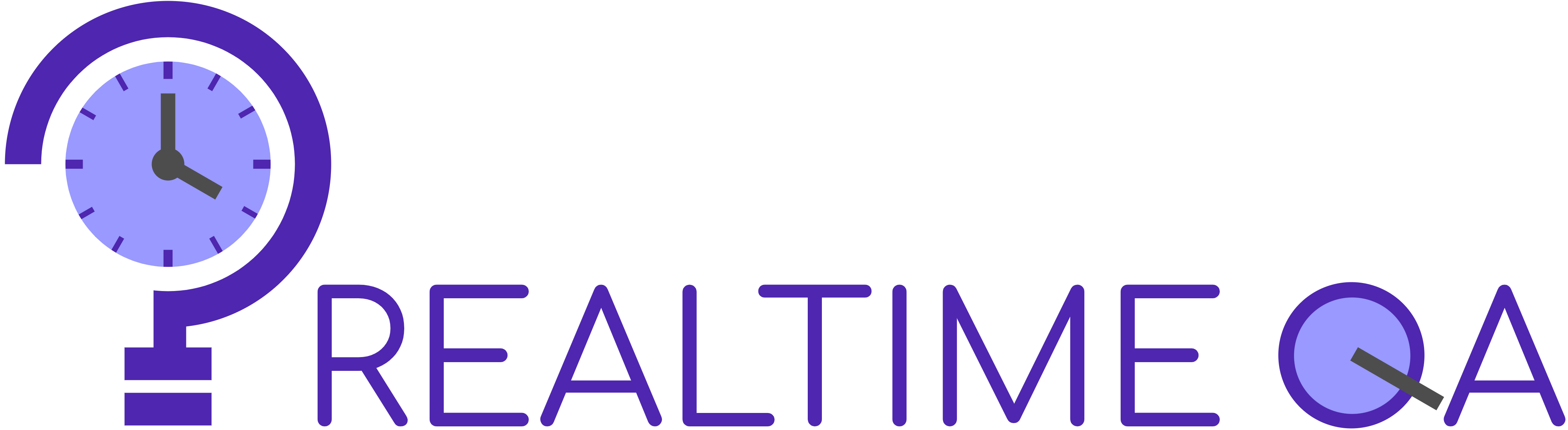}}}
\newcommand{\twitter}[0]{\raisebox{-0.0\height}{\includegraphics[width=.023\textwidth]{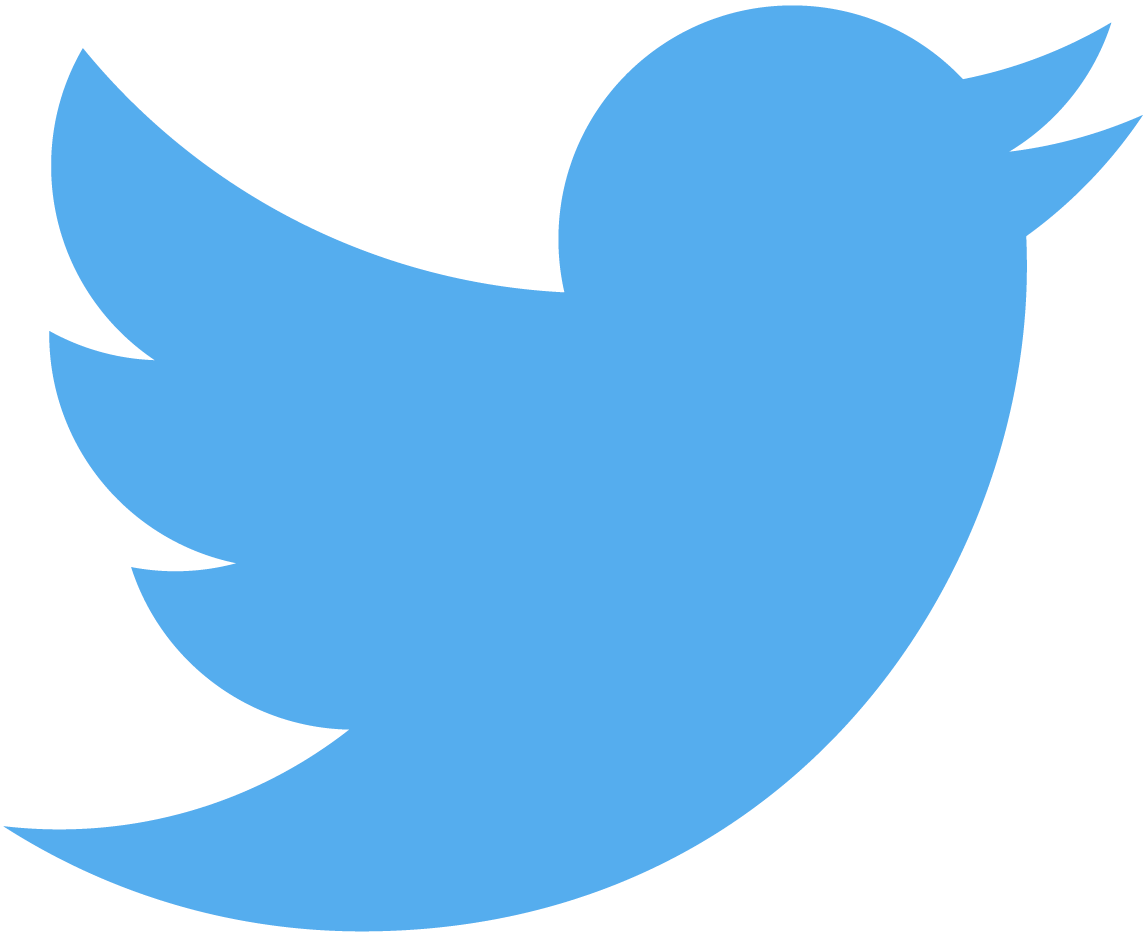}}}

\usepackage{color}
\usepackage{graphicx}
\usepackage[toc,page]{appendix}
\usepackage{makecell}
\usepackage{boldline}

\usepackage{algorithm}
\usepackage{colortbl}
\usepackage{tabstackengine}
\usepackage{tikz}
\usepackage{relsize}
\usepackage{microtype}
\usepackage{tablefootnote}
\usepackage{silence}
\usepackage{wrapfig}

\usepackage{tikz}
\usepackage{soul}

\usepackage{fdsymbol}

\usepackage{blindtext}
\usepackage{graphicx}
\usepackage{capt-of}
\usepackage{booktabs}
\usepackage{varwidth}
\usepackage{multirow}
\usepackage{xspace,mfirstuc,tabulary}

\newcolumntype{L}[1]{>{\raggedright\let\newline\\\arraybackslash\hspace{0pt}}m{#1}}
\newcolumntype{C}[1]{>{\centering\let\newline\\\arraybackslash\hspace{0pt}}m{#1}}

\definecolor{lightgray}{gray}{0.9}
\colorlet{soulgreen}{green!30}
\definecolor{red}{HTML}{FF0000}
\definecolor{blue}{HTML}{0000FF}
\definecolor{darkgreen}{HTML}{228B22}
\definecolor{dblue}{HTML}{007FFF}

\usepackage{booktabs,arydshln}

\makeatletter
\def\adl@drawiv#1#2#3{%
        \hskip.5\tabcolsep
        \xleaders#3{#2.5\@tempdimb #1{1}#2.5\@tempdimb}%
                #2\z@ plus1fil minus1fil\relax
        \hskip.5\tabcolsep}
\newcommand{\cdashlinelr}[1]{%
  \noalign{\vskip\aboverulesep
           \global\let\@dashdrawstore\adl@draw
           \global\let\adl@draw\adl@drawiv}
  \cdashline{#1}
  \noalign{\global\let\adl@draw\@dashdrawstore
           \vskip\belowrulesep}}
\makeatother

\usepackage[noend]{algpseudocode}
\algnewcommand{\parState}[1]{\State%
    \parbox[t]{\dimexpr\linewidth-\algmargin}{\strut\hangindent=\algorithmicindent \hangafter=1 #1\strut}}

\algrenewcommand\algorithmicindent{1.0em}%

\definecolor{magenta}{HTML}{F3DFF1}
\definecolor{red}{HTML}{FF0000}
\definecolor{hlgreen}{HTML}{D5E8D4}
\definecolor{figblue}{HTML}{DAE8FC}

\usepackage{algorithm}
\usepackage{colortbl}
\usepackage{tabstackengine}
\usepackage{tikz}
\usepackage{relsize}
\usepackage{microtype}
\definecolor{magenta}{HTML}{F3DFF1}
\definecolor{hlgreen}{HTML}{ccfcc4}
\definecolor{figblue}{HTML}{e7f2fe}

\DeclareRobustCommand{\greenbox}[1]{\setlength{\fboxsep}{1.0pt}\colorbox{green!25}{#1}}
\DeclareRobustCommand{\redbox}[1]{\setlength{\fboxsep}{1.0pt}\colorbox{red!15}{#1}}

\usepackage{pifont}
\newcommand{\xmark}{\textcolor{red}{\ding{55}}}
\newcommand{\cmark}{\textcolor{darkgreen}{\ding{51}}}

\usepackage{amsmath,amsfonts,bm}

\def\eqref#1{equation~\ref{#1}}

\def\1{\bm{1}}

\def\vy{{\mathbf{y}}}

\DeclareMathAlphabet{\mathsfit}{\encodingdefault}{\sfdefault}{m}{sl}
\SetMathAlphabet{\mathsfit}{bold}{\encodingdefault}{\sfdefault}{bx}{n}

\title{\realtime: What's the Answer Right Now?}
\author{%
    Jungo Kasai$^{\heartsuit}$\thanks{\ \ Work was done during JK's internship at AI2.}  
\quad
\textbf{Keisuke Sakaguchi}$^{\clubsuit\square}$\footnotemark[1]
\quad
\textbf{Yoichi Takahashi}$^{\clubsuit}$
\quad 
\textbf{Ronan Le Bras}$^{\diamondsuit}$
\\
\textbf{Akari Asai}$^{\spadesuit}$
\quad
\textbf{Xinyan Velocity Yu}$^{\varheartsuit}$
\quad
\textbf{Dragomir Radev}$^{\vardiamondsuit}$
\\
\textbf{Noah A.\ Smith}$^{\diamondsuit\spadesuit}$
\quad
\textbf{Yejin Choi}$^{\diamondsuit\spadesuit}$
\quad 
\textbf{Kentaro Inui}$^{\triangle\clubsuit\square}$
\\
$^{\heartsuit}$Toyota Technological Institute at Chicago
\quad
     $^{\clubsuit}$Tohoku University
     \quad
     $^{\square}$RIKEN
     \quad
$^{\diamondsuit}$Allen Institute for AI
\\
$^{\spadesuit}$University of Washington
\quad
    $^{\varheartsuit}$University of Southern California
\quad
    $^{\vardiamondsuit}$Yale University
     \quad
     $^{\triangle}$MBZUAI
     \\
     \\
    \logo
    \quad 
    \texttt{realtimeqa.nlp@gmail.com}
    \quad 
    \twitter\texttt{@realtimeqa}
    \\
}

\begin{document}

\maketitle
\begin{abstract} 
We introduce \realtime, a dynamic question answering (QA) platform that announces questions and evaluates systems on a regular basis (weekly in this version).
\realtime inquires about the \emph{current} world, and QA systems need to answer questions about novel events or information.
It therefore challenges static, conventional assumptions in open-domain QA datasets and pursues instantaneous applications. 
We build strong baseline models upon large pretrained language models, including GPT-3 and T5.
Our benchmark is an ongoing effort, and this paper presents real-time evaluation results over the past year.
Our experimental results show that GPT-3 can often properly update its generation results, based on newly-retrieved documents, highlighting the importance of up-to-date information retrieval.
Nonetheless, we find that GPT-3 tends to return \emph{outdated} answers when retrieved documents do not provide sufficient information to find an answer.
This suggests an important avenue for future research: can an open-domain QA system identify such unanswerable cases and communicate with the user or even the retrieval module to modify the retrieval results?
We hope that \realtime will spur progress in instantaneous applications of question answering and beyond.\footnote{\url{https://realtimeqa.github.io/}.}
\end{abstract}
\section{Introduction}
\begin{wrapfigure}{r}{0.47\textwidth}
\vspace{-1cm}
\centering    \includegraphics[width=0.46\textwidth]{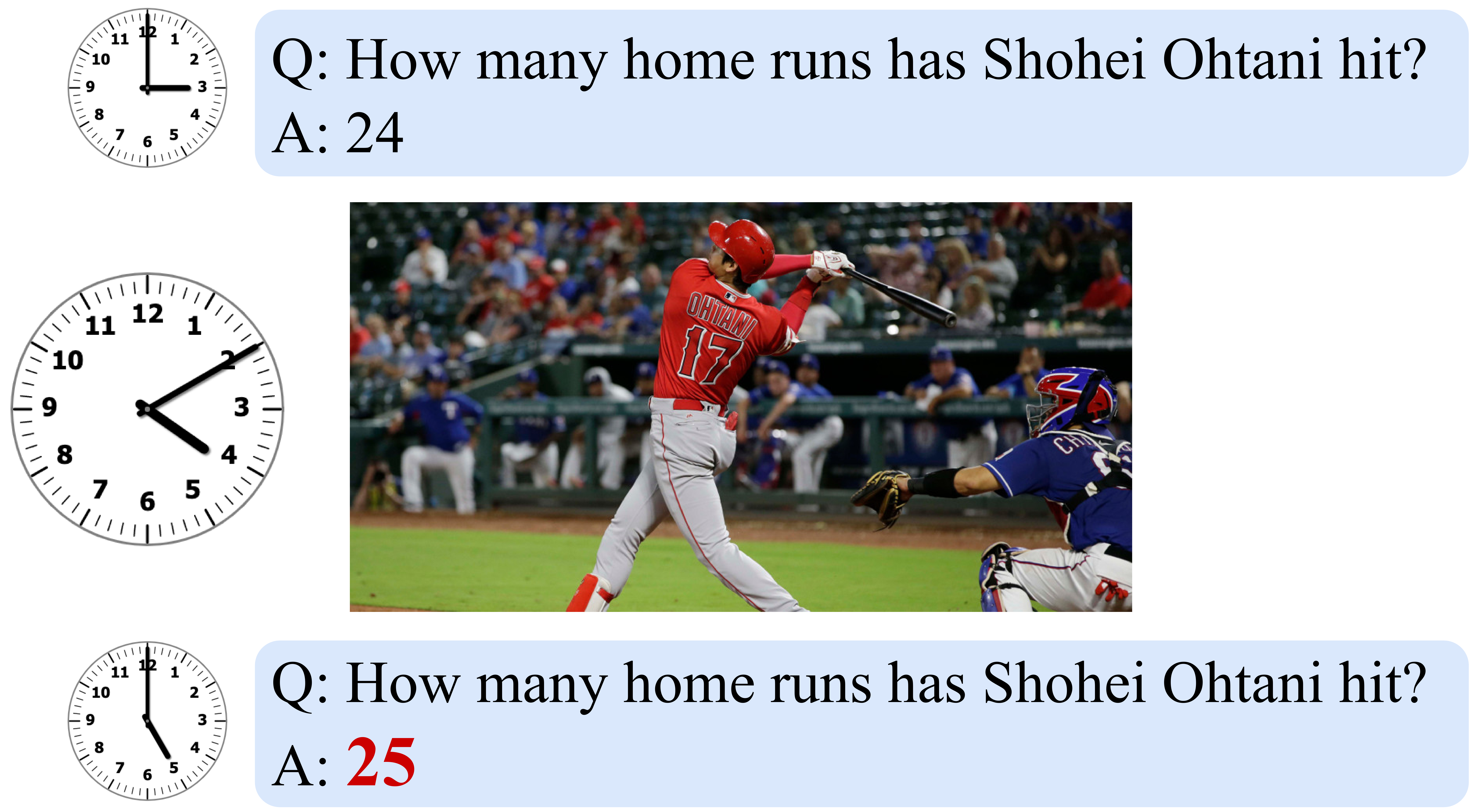}
\caption{\realtime establishes a framework to benchmark question answering at the present time: answers (e.g., the number of Shohei Ohtani's home runs) change in real time. Source: \url{https://thecomeback.com/mlb/shohei-ohtani-home-runs-tommy-john.html}.}
\vspace{-1.0cm}
\label{fig:real_time}
\end{wrapfigure}

\textit{How many home runs has Shohei Ohtani
hit so far this season?}
A user of a question answering (QA) system might ask such time-sensitive questions and seek out answers in \textit{real time}.
Widely-used evaluation benchmarks of QA systems, however, implicitly assume that answers are static regardless of the time of inquiry.
Several recent works \citep{tempquestions18,time-sensitive,zhang-choi-2021-situatedqa,streamingqa2022} challenged this assumption and proposed QA datasets that specify the temporal context (e.g., \textit{who was the President of the U.S. in 1940?}).
We extend these recent efforts on time-sensitive QA to fulfill real-time, more instantaneous information needs from users:
we establish a dynamic benchmark based on newly-published news articles---\realtime---and provide a regularly-updated (weekly in the current version) evaluation platform for the research community.


We develop an annotation framework (\S\ref{section:framework}) and a benchmarking timeline for real-time QA system submissions.
Every week, \realtime retrieves news articles and human-written, multiple-choice questions from news websites (CNN, THE WEEK, and USA Today), covering a wide range of topics, including politics, business, sports, and entertainment.
We upload these data, as well as our baseline results, to our website, and any model submission can be evaluated until the next set of questions is posted.
This dynamic scheme contrasts with the well-established QA annotations \citep{chen2017reading,chen-yih-2020-open} that are performed only \emph{once} with information available at the time.
Such annotations are effective for factoid \citep{berant-etal-2013-semantic,teaching_machines,rajpurkar-etal-2016-squad,joshi-etal-2017-triviaqa} or commonsense questions \citep{zellers-etal-2018-swag,zellers-etal-2019-hellaswag,talmor-etal-2019-commonsenseqa,winogrande}, but not the real-time information needs that are our target.
\begin{wrapfigure}{r}{0.6\textwidth}
\centering
    \includegraphics[width=0.6\textwidth]{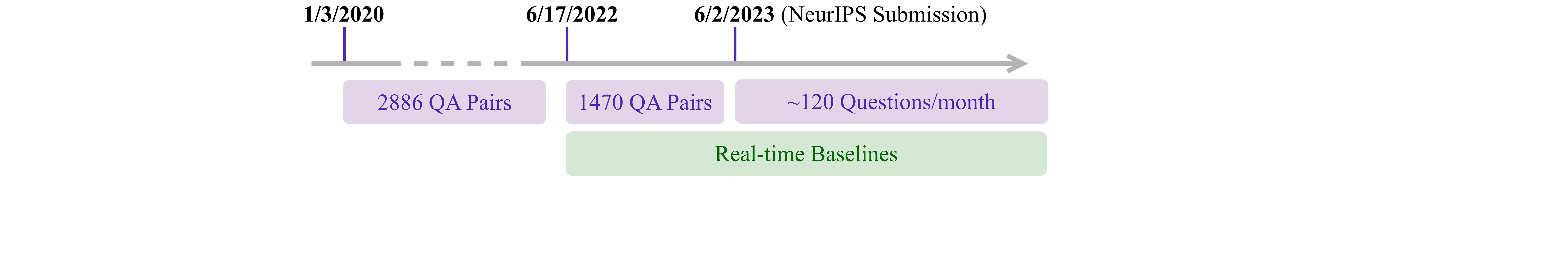}
\caption{\realtime data statistics as of June 2, 2023. We started our real-time baselines on June 17, 2022 (\S\ref{section:baselines}).
We also provide past 2,886 QA pairs that can be used by model developers (e.g., finetuning).}
\label{fig:data_stats}
\end{wrapfigure}

We present two classes of real-time baseline systems that are built on strong, recent models (GPT-3: \citealp{gpt3}; T5: \citealp{2020t5}; BART: \citealp{lewis-etal-2020-bart}): open-book and closed-book QA models.
We present a prompting method to use GPT-3 for open-domain QA.
The former class uses an external knowledge source, such as Wikipedia \citep{min2019knowledge,guu2020realm,rag2020,izacard2021leveraging} or news articles.
The latter class of closed-book models directly outputs an answer to each question.
By design, these closed-book baselines have no access to information more recent than the time of pretraining or finetuning, thereby helping us understand the degree to which real-time information is truly necessary.
Notably, a small number of questions in \realtime ($\sim$12$\%$) do not strictly require recent information; for example, Shohei Ohtani hits a home run today, leading one to ask where he was born.
This is consistent with information-seeking, naturally-occurring scenarios that we target in this work, as seen in \citet{clark-etal-2020-tydi}.
Most users of a QA system do not exclusively ask time-sensitive questions, even though these questions may be stimulated by current events; QA systems \emph{should} aim to address these questions as well.
\begin{figure*}[h]
\centering
    \includegraphics[width=0.90\textwidth]{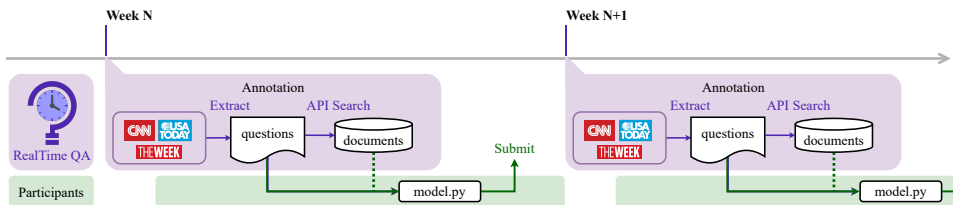}
\caption{\realtime annotation framework and submission workflow.
At 3 am GMT on every Saturday, we extract questions from news websites and post them on our website.
We immediately run API search for these questions (Google custom search) and share the results as a document pool.
The use of this document pool is optional (indicated by a dashed line); participants are allowed to retrieve evidence documents by themselves.
All evaluations are done on our website, and the submission window closes when the next set of questions is announced.}
\label{fig:framework}
\end{figure*}
\begin{figure*}[t]
\centering
    \includegraphics[width=0.85\textwidth]{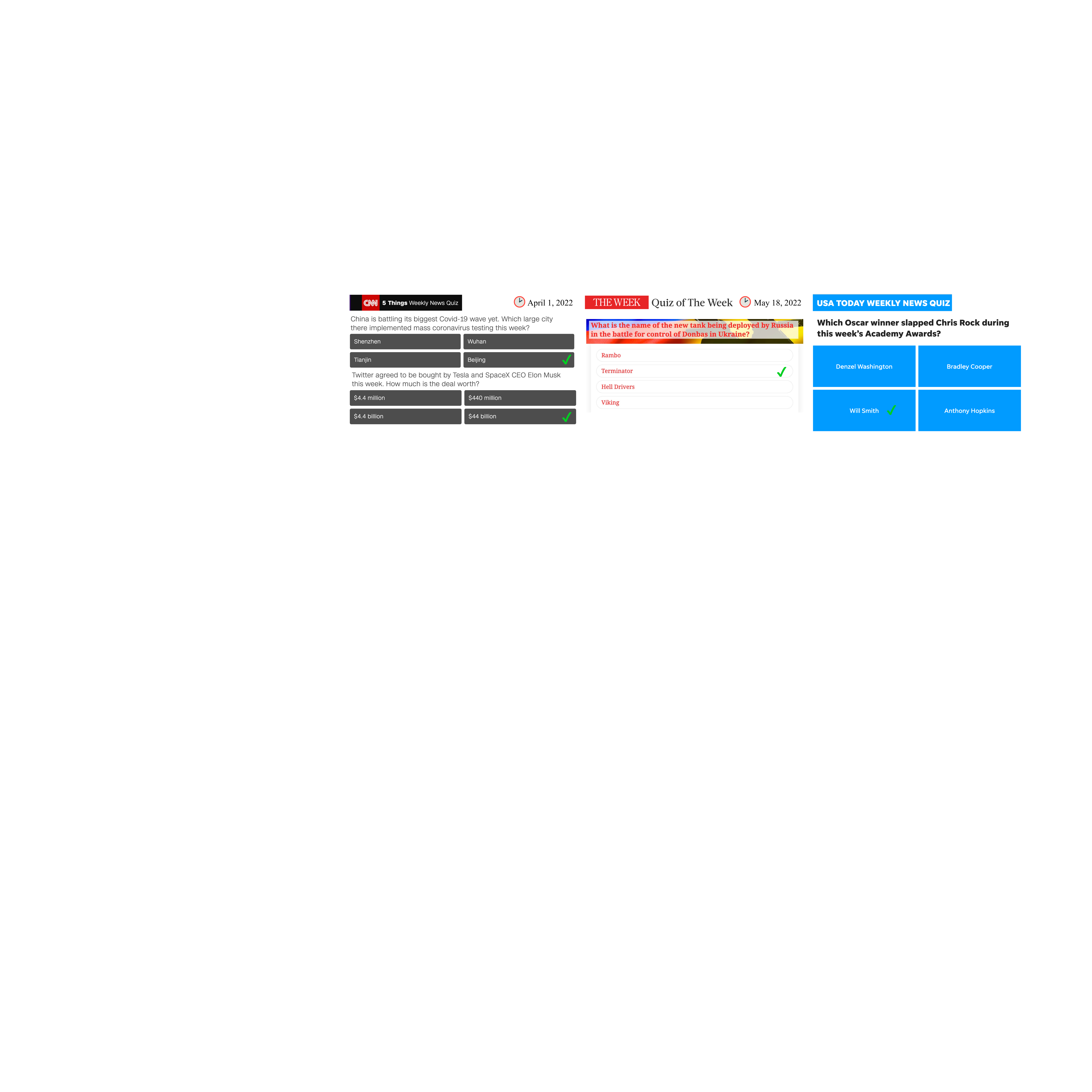}
\caption{Examples of weekly quizzes from CNN and THE WEEK that are extracted during annotations of \realtime. They span diverse genres, including politics, business, and entertainment.}
\label{fig:realtime_example}
\end{figure*}

We evaluate six baselines both in multiple-choice and generation settings \textit{in real time} and report the results over the period of June 17 through June 2, 2023.
These evaluation data resulted in a total of 1,470 QA pairs (Fig.\ \ref{fig:data_stats}). 
Further, we provide 2,886 QA pairs that are collected in the same way but preceded our real-time evaluations.
These can be used in later work for model development (e.g., finetuning).
Our results show that an open-book GPT-3 model augmented with up-to-date text retrieval substantially outperforms closed-book baselines, as well as open-book models with retrieval from a past Wikipedia dump \citep{rag2020}.
This result illustrates that large language models can adjust their knowledge, based on the retrieved passages (\S\ref{section:experiments}).
Nonetheless, we find that they still struggle, especially when the multiple choices include uncertainty (e.g., ``none of the above'').
Most of the errors originate from retrieval, rather than reading comprehension.
The \realtime benchmark, therefore, highlights the importance of fast, up-to-date text retrieval \citep{seo-etal-2019-real} to better serve instantaneous information needs.
We share all data and code to reproduce our baselines so that follow-up work can build upon our first attempts to tackle this task.


\realtime can also serve as an important step toward much-needed, broader, real-time applications of NLP.
For example, a QA system with timely updates can improve emergency management of natural disasters \citep{imran2013practical,imran2015,imran2016lrec,nguyen_rapid} or pandemics (e.g., COVID-19; \citealp{wang-etal-2020-cord,lee-etal-2020-answering,moller-etal-2020-covid,Alzubi2021COBERTCQ}).
With the advent of online news, prior work developed automated systems that regularly retrieve and summarize news articles from the Internet \citep{allan01,newsisessence,tracking_newsblaster,mckeown-etal-2003-columbias,evans-etal-2004-columbia}.
Models developed for the \realtime task can be further enhanced with such retrieval/summarization systems.
We hope that our \realtime interface and baseline models will serve as a useful platform for research and real-world applications.

\section{\realtime Framework}
\label{section:framework}
Our current version announces questions every week, based on news articles published within the past week.
Here we establish the workflow (\S\ref{section:timeline}) and the framework for annotations (\S\ref{section:annotation}) and evaluations (\S\ref{section:evaluation}).
We then discuss our built-in baselines (\S\ref{section:baselines}) that are continually evaluated every week.
Our user interface and more detailed statistics (e.g., genres and answer types) are available in Appendices \ref{appendix:interface} and \ref{appendix:stats}.

\subsection{Workflow}
\label{section:timeline}
Fig.\ \ref{fig:framework} depicts the \realtime workflow for each week.
We announce $\sim$30 multiple-choice questions at 3 am GMT every Saturday.
We internally run API search (Google custom search, GCS) for these questions and share a set of documents (mostly news articles) with the URLs that are available at that time.
Participants run their model on these questions, optionally using the documents from our API search as a knowledge source (indicated as dashed lines in Fig.\ \ref{fig:framework}).
While we provide our document set to lower barriers to submission, \textbf{participants are also allowed to create and use knowledge sources by themselves} (e.g., custom retrieval models or other external APIs such as Twitter API).
System submissions are shared on our website with their performance and submission time.
The submission window closes when the new set of questions is announced the next week.

Note that fair, \textit{retroactive} comparisons of systems are also possible, as long as they use data available when the submission window was still open.
For instance, participants might be interested in evaluating their model against a past submission on the Week N questions.
In this case, they can do so by ensuring that their system only relies on data up to Week N and simulating how their system \textit{would have performed} at that time.
Our platform still focuses on real-time evaluations and encourages every participant to submit real-time results to better reflect real-world applications.

\subsection{Annotation}
\label{section:annotation}
\noindent\textbf{Question Extraction} \ 
The authors of this paper perform weekly annotations in a human-in-the-loop way.
We first find web pages for ``weekly quizzes'' from three news websites: CNN (US-based), USA Today, and The WEEK (UK-based).\footnote{Fair use is allowed under Section 107 of the Copyright Act in the U.S.: \url{https://www.copyright.gov/title17/92chap1.html\#107}.}
Shown in Fig.\ \ref{fig:realtime_example} are examples that span politics and business genres.
We then execute our extraction script to collect multiple-choice questions.
Each of these three websites posts $\sim$10 questions per week, resulting in $\sim$120 questions in total every month.
Weekly quizzes are also available from the New York Times and ABC Australia, but they are not included in the current version, due to issues with automatic extraction or a paid subscription system.

\noindent\textbf{API Search} \
Using each of these questions as a retrieval query, we run Google custom search\footnote{\url{https://programmablesearchengine.google.com/}.} to collect the top-10 documents from the web.
The retrieval target is all articles from CNN, USA Today, and THE WEEK.
We then parse every document using the \texttt{newspaper3k} package\footnote{\url{https://github.com/codelucas/newspaper}.} and store the text as well as metadata, such as the publication date and author name.
In some rare cases, articles from the search get taken down,
in which case we disregard them.
This indeed illustrates a unique challenge of real-time applications with constantly-changing, dynamic information.

\subsection{Evaluation}
\label{section:evaluation}
\noindent\textbf{Multiple Choice}  \
Since \realtime is a multiple-choice question dataset, we can simply measure performance by accuracy.
We also explored a NOTA (none of the above) setting: one of the original choices is randomly replaced with ``none of the above,'' thereby helping prevent models from exploiting heuristics \citep{rajpurkar-etal-2018-know}.
As expected, the NOTA setting resulted in performance degradation across the board (\S\ref{section:results}).
NOTA choices can be found in other multiple-choice QA or reading comprehension datasets \citep{richardson-etal-2013-mctest,lai-etal-2017-race}.

\noindent\textbf{Generation}  \
We also experiment with a generation setting where no choices are given, to better reflect real-world applications.
Under this setting, we evaluate performance with exact matching and token-based F1 scores, following the standard practice in question answering \citep{rajpurkar-etal-2016-squad}.

\noindent\textbf{Human Performance} \
We randomly sampled 10 weeks from June 17, 2022 through January 13, 2023 (300 questions in total), and the authors of this paper answered multiple-choice questions using Google search.
This resulted in the accuracy of $96.7\%$.
Most questions in \realtime are straightforward (e.g., single-hop questions) and a human with Internet access can easily answer them.\footnote{In fact, USA Today has a record of human top scorers every week, and they all get perfect scores. E.g., \url{https://www.usatoday.com/storytelling/quiz/news-quiz/2022-07-01/}.}
For the sustainability of the dynamic benchmark, we do not provide an estimate of human performance on a regular basis.

\subsection{Real-time Baselines}
\label{section:baselines}
\realtime executes six baselines in real time that are based on strong pretrained models: four open-book and two closed-book models.
These six models are evaluated and made publicly available when weekly questions are announced.
Any submission to \realtime is compared against them.
Participants can also build their model upon our baselines. See Appendix \S\ref{appendix:baseline_configs} for more detail.

\subsubsection{Open-book QA Models}
\label{section:open-book}
Open-book QA models follow a two-step pipeline: \textbf{document retrieval} that finds evidence documents from an external knowledge source (e.g., Wikipedia) and \textbf{answer prediction} (or reading comprehension) that outputs an answer conditioned on the question and evidence documents.
For either step, we experiment with two variants, resulting in a total of four configurations.
Open-book systems have the advantage of being capable of updating the external knowledge source at test time \citep{rag2020}.
This property is particularly crucial for questions in \realtime that inquire about information at the present time.

\noindent\textbf{Document Retrieval} \
For the retrieval step, we experiment with two configurations: top-5 Wikipedia documents from dense passage retrieval (\textbf{DPR}; \citealp{karpukhin-etal-2020-dense}) and top-5 news articles from \textbf{GCS} (\S\ref{section:annotation}).
In DPR, English Wikipedia articles from the December 2018 dump are segmented into 100-word documents \citep{wang-etal-2019-multi}.
DPR encodes the question and every document into 768-dimensional vectors; it then computes the inner product to obtain a matching score and selects documents with top-5 matching scores.
We use the BERT-based model \citep{devlins2019bert}, finetuned on the Natural Questions dataset \citep{kwiatkowski2019natural} from the Hugging Face Transformers library \citep{wolf-etal-2020-transformers}.
GCS uses an external API, and we found that it sometimes returned fewer than five documents ($\sim$10\% of the time); in this case, we add top documents from DPR to create a top-5 document set.

\noindent\textbf{Answer Prediction} \
We explore two methods for answer prediction, conditioned on the question and the corresponding retrieved text: retrieval-augmented generation (\textbf{RAG}; \citealp{rag2020}) and a prompting method with \textbf{GPT-3} (text-davinci-002; \citealp{gpt3}).
In the multiple-choice setting, we compute the log probability of every choice and normalize it by the generation sequence length.
We then select the choice with the best score.
For the generation setting, we simply perform text decoding.

For the \textbf{RAG} baseline, we use the BART-based \citep{lewis-etal-2020-bart} RAG-sequence model, again finetuned on Natural Questions from the Transformers library.
This model predicts the answer sequence $\vy$ autoregressively from left to right while marginalizing over the set of top-5 retrieved documents ($\mathcal{Z}$):
$ P\left(\vy\right) = \sum_{z \in \mathcal{Z}} P \left(z\right) \prod_{t=1}^{|\vy|} P \left( y_t | z, \vy_{\leq t}\right) $.
Here $P(z)$ is given by the matching score from the retrieval step.\footnote{Unlike DPR, GCS does not provide matching scores. We treat top-5 documents with equal probabilities.}
In the equation, the conditioned-upon question is suppressed for brevity.

\begin{wrapfigure}{R}{0.55\textwidth}
\centering
\vspace{-1.2cm}
\includegraphics[width=0.52\textwidth]{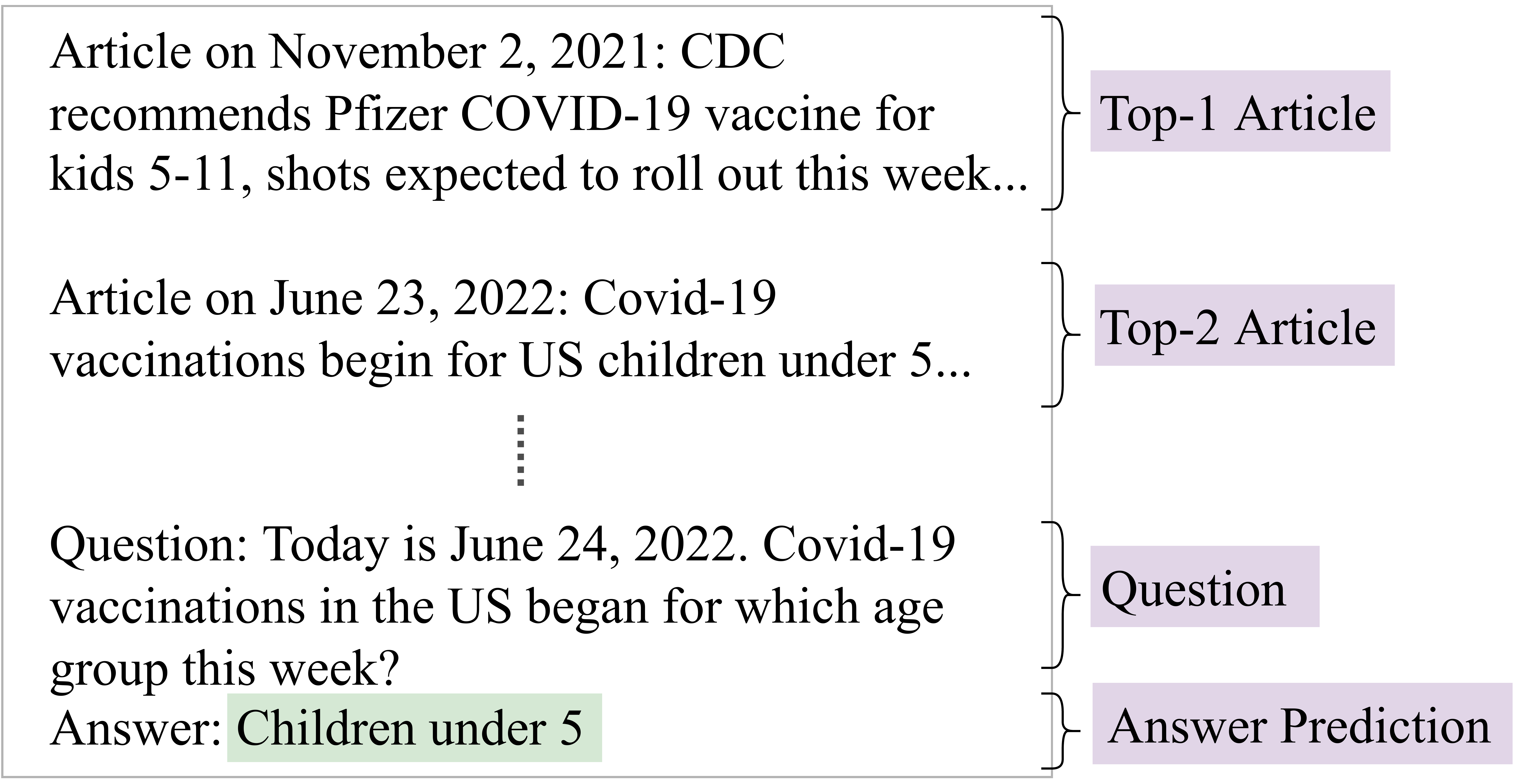}
\caption{Example prompt for answer generation with the open-book GPT-3 baseline. For the closed-book GPT-3 baseline, the top-5 articles are not given. We perform ablation studies on the date information (\S\ref{section:analysis}).}
\label{fig:gpt3_prompt}
\vspace{-0.5cm}
\end{wrapfigure}

We propose a straightforward \textbf{GPT-3} prompting method with temporal contexts (Fig.\ \ref{fig:gpt3_prompt}).\footnote{See \citet{internet_gpt3} for other prompt templates.}
We prepend to every question the title and the first two paragraphs of the top-5 articles from the document retrieval step.\footnote{This substantially reduces the inference cost. They contain most of the key information in each article.}
The publication date is inserted, using the metadata of each retrieved article (e.g., ``Article on November 2, 2021'' in Fig.\ \ref{fig:gpt3_prompt}).
For Wikipedia passages retrieved by DPR, we prepend ``December 31, 2018,'' based on the Wikipedia dump date \citep{karpukhin-etal-2020-dense}.
Our ablation studies on date insertion will show that the open-book GPT-3 system benefits from specifying the dates of the question and the retrieved articles to some extent (\S\ref{section:analysis}).

\subsubsection{Closed-book QA Models}
Closed-book QA models directly answer questions without access to external knowledge.
They have proven competitive with open-book models on some QA datasets \citep{roberts-etal-2020-much,guu2020realm}.
Since these models are trained/finetuned on the data available at that time, they cannot address questions about new events or updated information.
Nonetheless, some of the real-time information needs do not necessarily require up-to-date information.
Indeed, \realtime contains a small portion of such questions ($\sim$10$\%$).
For instance, 
\textit{Microsoft retired its Internet Explorer browser this week. What year did it debut?}
Such questions are triggered by a new event but inquire about facts in the past that have not changed recently.
Most users of a QA system do not exclusively raise time-sensitive questions, and QA systems should aim to address these questions as well.
Closed-book baselines thus quantify the degree to which up-to-date information is necessary to answer questions in \realtime. 
We use the following two strong methods for closed-book QA.

\noindent\textbf{Finetuning Method} \
We use the T5 model (T5-11B; \citealp{2020t5}) finetuned on the Natural Questions data, again from the Transformers library.
Following the open-book baseline, we select the choice with the maximum average log score in the multiple-choice setting.

\noindent\textbf{Prompting Method} \
Similar to the open-book baselines (\S\ref{section:open-book}), we apply a prompting method to GPT-3 (text-davinci-002).
We use the same prompt as Fig.\ \ref{fig:gpt3_prompt}, except that no articles are inserted before the question. 
Again following the open-book baselines, we select the choice with the maximum average log score in the multiple-choice setting.

\section{Experiments and Analysis}
\label{section:experiments}
\label{section:results}
We started our real-time experiments on June 17 2022, spanning a year as of June 2 2023 (1470 questions in total).
We will continue our weekly annotations, but here we report our experimental and analysis results so far and give guidance to future participants.

\begin{table}[h!]
\parbox{.48\linewidth}{
\centering
\caption{Results from the past year (from June 17, 2022 through June 2, 2023). GCS: Google custom search; DPR: dense passage retrieval \citep{karpukhin-etal-2020-dense}; RAG: retrieval-augmented generation \citep{rag2020}.}
 \addtolength{\tabcolsep}{-3pt}  
\small
\begin{tabular}{@{} l @{\hskip -1.0em} cc m{0.1em} cc  m{0.1em}  cc @{} }
\toprule[.1em]

\multicolumn{3}{c}{\textbf{Real-time Baselines}}
&&
\multicolumn{2}{c}{\textbf{Multi-choice}}
&&
\multicolumn{2}{c}{\textbf{Generation}}
\\
\cmidrule(lr){1-3}
\cmidrule(lr){5-6}
\cmidrule(lr){8-9}
& \textbf{Retrieve}
& \textbf{Predict}
&& \textbf{Orig.}
&\textbf{NOTA}
&&\textbf{EM}
& \textbf{F1}
\\

\midrule[.1em]
\multirow{4}{*}{Open}
&DPR
&RAG
&& 27.4
& 24.8
&& 2.4
& 4.1
\\

&DPR
&GPT-3
&&  43.9
& 35.8
&& 13.3
& 19.7
\\

&GCS
&RAG
&& 46.9
& 37.9
&&  17.5
& 22.1
\\

&GCS
&GPT-3
&& \textbf{66.5}
& \textbf{58.4}
&&  \textbf{34.6}
& \textbf{45.3}
\\

\midrule[.05em]
\multirow{2}{*}{Closed}
& ---
& T5
&& 39.1
& 35.3
&& 9.7
& 14.7
\\

& ---
& GPT-3
&& 44.9
& 34.1
&& 15.3
& 22.3
\\

%



\bottomrule[.1em]
\end{tabular}
\label{tab:main_results}
}
\hfill
\parbox{.48\linewidth}{
\centering
\caption{Ablation studies on date insertion in the prompt for the open-book (Google custom search; GCS) and close-book GPT-3 baselines. All results are averaged over the first six weeks: June 17 through July 22, 2022.}
 \addtolength{\tabcolsep}{-3pt}  
\small
\begin{tabular}{@{} l @{\hskip -1.0em} cc m{0.1em} cc  m{0.1em}  cc @{} }
\toprule[.1em]

\multicolumn{3}{c}{\textbf{Date Insert}}
&&
\multicolumn{2}{c}{\textbf{Multi-choice}}
&&
\multicolumn{2}{c}{\textbf{Generation}}
\\
\cmidrule(lr){1-3}
\cmidrule(lr){5-6}
\cmidrule(lr){8-9}
& \textbf{Articles}
& \textbf{Qs}

&& \textbf{Orig.}
&\textbf{NOTA}
&&\textbf{EM}
& \textbf{F1}
\\

\midrule[.1em]
\multirow{4}{*}{
Open}
& \cmark
& \cmark
&&  \textbf{69.3}
&  59.8
&&  \textbf{28.7}
&  \textbf{39.4}
\\

&\cmark
&\xmark
&&  66.5
& \textbf{62.6}
&& 24.7 
& 36.3 
\\

& \xmark
& \cmark
&& 67.0
&  57.5 
&& 28.1
& 38.2
\\

& \xmark
& \xmark
&& 65.9
& 61.5 
&& \textbf{28.7}
& 38.3
\\

\midrule[.05em]
\multirow{2}{*}{Closed}
& ---
& \cmark
&& 39.7
& 31.3 
&& 7.3 
& 15.2
\\

& ---
& \xmark
&& 45.8
& 38.5 
&& 9.0
& 15.9 
\\

%



\bottomrule[.1em]
\end{tabular}
\label{tab:date_insertion}
}
\end{table}

\subsection{Results}
Seen in Table \ref{tab:main_results} are the results from the past year.
In all three settings (original/NOTA multiple choice and generation), GPT-3 with Google custom search (GCS) retrieval achieves the best performance.
In particular, GPT-3 with GCS substantially outperforms both closed-book GPT-3 and GPT-3 with DPR (from a December 2018 Wikipedia dump): e.g., 34.6 vs.\ 15.3/13.3 in generation exact matching.
This suggests that GPT-3 is able to answer questions based on the given prompt, rather than relying on past information from pretraining.
Nevertheless, we still see a large performance drop of all baselines from the original multiple-choice setting to NOTA (``none of the above''): e.g., 58.4 vs.\ 66.5 for GPT-3 with GCS retrieval.
Future work can further improve GPT-3's ability of reading comprehension, especially regarding answer uncertainty.

\subsection{Analysis and Ablations}
\label{section:analysis}



\noindent\textbf{Date Insertion for Prompting} \
Our prompt for the GPT-3 baselines prepends date information both to the articles and question (Fig.\ \ref{fig:gpt3_prompt}).
Table \ref{tab:date_insertion} shows results from ablation studies on date insertion for the open-book (GPT-3 with Google custom search) and closed-book GPT-3 models.
Temporal specification almost always helps the open-book GPT-3 model. 
Interestingly, it hurts the performance of the closed-book model, perhaps because the specified date is generally unseen during pretraining and the prompt becomes ``out-of-domain.''

\noindent\textbf{Error Breakdown} \
We conducted a manual error  analysis of the results so far.
In particular, we categorized answers from the best generation model (open-book GPT-3 with GCS) into three categories: correct, retrieval error, and reading comprehension error.
For the questions from the first six weeks, the breakdown was the following: correct (52\%), retrieval error (34\%), and reading comprehension error (13\%).
This suggests that the key to instantaneous applications of question answering is \textbf{accurate, up-to-date information retrieval}.

\begin{wrapfigure}{R}{0.5\textwidth}
\centering
\includegraphics[width=0.45\textwidth]{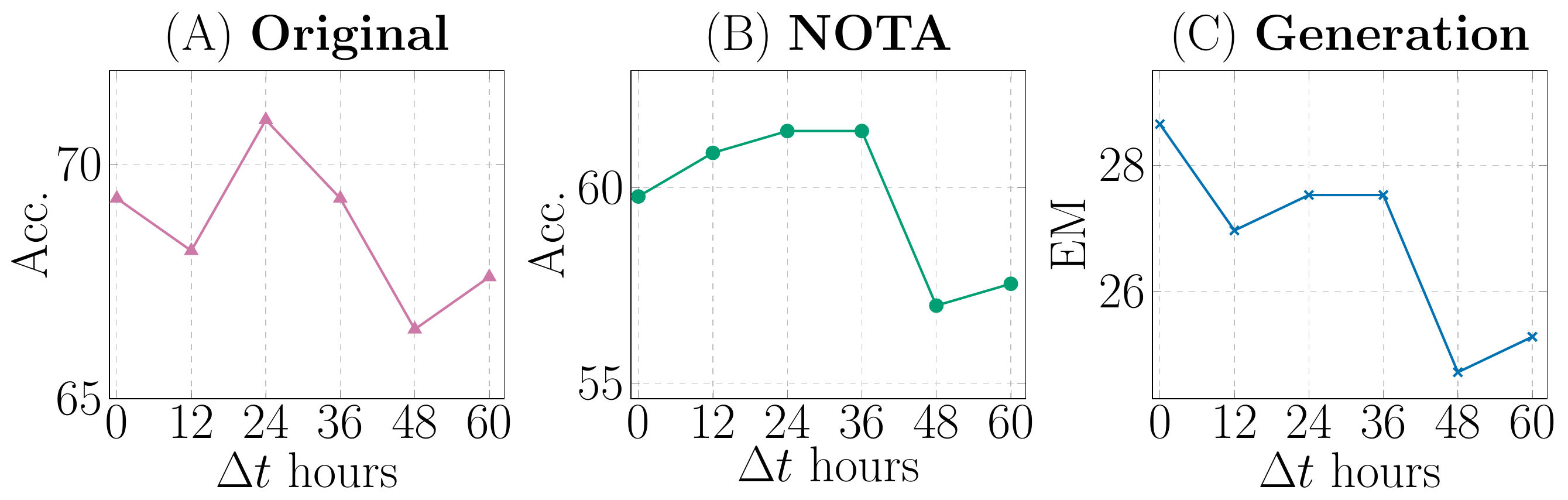}
\caption{Performance vs.\ submission time (hours after the announcement of questions, 3 am GMT on Saturday) over the three evaluation settings (A: original multiple choice; B: none of the above; C: generation).
All results are from open-book GPT-3 with Google custom search (GCS) and averaged over the questions from June 17, 2022 through July 22, 2022.
$\Delta t\!=\!0$ for all of our six real-time baselines by default.
}
\label{fig:submission_time}
\end{wrapfigure}

\noindent\textbf{Performance vs.\ Submission Time}
Fig.\ \ref{fig:submission_time} plots the performance of the open-book GPT-3 baseline with Google custom search (GCS) over varying submission (i.e., GCS retrieval) time.
All results are averaged over the questions between June 17 and July 22, 2022.
We see a consistent pattern: the performance remains high (or improves) up to around 24 hours after the announcement but substantially degrades later.
While the performance can improve when GCS starts to retrieve more recent articles, it eventually suffers from temporal gaps.
Our website provides the submission time of every system as well as its performance.

\noindent\textbf{Examples} \
Table \ref{tab:examples} shows some examples that compare the closed-book and open-book GPT-3 models.
The first three examples illustrate that GPT-3 can correctly update its answer based on the retrieved documents across diverse genres: natural disasters, the COVID-19 pandemic, and entertainment. 
The last three cases, on the other hand, demonstrate a critical limitation of current large language models in temporal understanding: \textbf{the retrieved documents do not suffice to answer the questions due to a temporal gap, and GPT-3 still generates an outdated answer}.
Ideally, GPT-3 should inform the user or even the retrieval module that it does not have enough evidence to answer the question. This way, the retrieval module can expand its search, or the user can consult other resources. 

Note that it is possible to limit the retrieval target to recent articles,\footnote{Indeed, GCS has a \emph{paid} version with a date range feature that filters retrieval results by date.} but there are potential failure modes.
Firstly, some questions in \realtime inquire about the past, and models can benefit from older articles when answering such questions.
Further, the appropriate date range for retrieval varies from question to question in real-world applications; some questions inquire about this year, while others about this week.
We thus do not implement such filtering for the current real-time baselines.

\begin{table*}
\small
\caption{Examples that compare closed-book and open-book GPT-3 answers with top-5 articles from Google custom search (GCS) retrieval. As in the first three examples, GPT-3 can adjust its answer based on newly-retrieved documents. When the retrieved documents are \emph{outdated} or unrelated, however, GPT-3 ignores the temporal gap and yields an outdated answer.}
\addtolength{\tabcolsep}{-3pt}
\begin{center}
\scalebox{0.9}{
\begin{tabular}{ L{6cm} L{4.5cm} L{4.5cm}}
\toprule
\textbf{Question} & \textbf{Retrieved Documents (Top-5)}  \\

\midrule[.04em]

Historic rainfall led to flooding, mudslides and visitor evacuations at which national park?

\par
\textbf{Date}: June 17, 2022
\par
\textbf{Answer}: 
Yellowstone National Park
\par
\textbf{Closed GPT-3}: \redbox{Yosemite National Park}
\par
\textbf{Open GPT-3}: \greenbox{Yellowstone National Park}

&
\textbf{June 14, 2022}
\par
\greenbox{Yellowstone National Park} flooding `still raging'...
\par
\textbf{June 13, 2022}
\par
\greenbox{Yellowstone National Park} closes entrances, evacuates visitors amid `unprecedented' rainfall...
\par
\textbf{June 15, 2022}
\par
Dozens evacuated as unprecedented flooding forces \greenbox{Yellowstone} \greenbox{National Park} to close...
&
\textbf{June 15, 2022}
\par
\greenbox{Yellowstone} still closed as flooding recedes and thousands evacuate...
\par
\textbf{June 14, 2022}
\par
Home swept away as \greenbox{Yellowstone National Park} is hit by major floods and mudslides...
\\

\midrule[.02em]
Covid-19 vaccinations in the US began for which age group this week?
\par
\textbf{Date}: June 24, 2022
\par
\textbf{Answer}: 
Children under 5
\par
\textbf{Closed GPT-3}: \redbox{18 and up}
\par
\textbf{Open GPT-3}: \greenbox{Children under 5}

&
\textbf{November 2, 2021}
\par
CDC recommends Pfizer COVID-19 vaccine \redbox{for kids 5-11}, shots expected to roll out this week...
\par
\textbf{June 23, 2022}
\par
Covid-19 vaccinations begin \greenbox{for US children under 5}...

&
\textbf{July 22, 2021}
\par
Biden says \redbox{kids under 12} could be eligible for COVID vaccines in weeks...
\par
\textbf{November 10, 2021}
\par
COVID-19 cases on the rise again in Iowa...
\par
\textbf{November 1, 2021}
\par
Everything to know about COVID-19 vaccine and children...
\\

\midrule[.02em]

Which wildly popular show was recently green lit for a new season?
\par
\textbf{Date}: June 17, 2022
\par
\textbf{Answer}: Squid Game
\par
\textbf{Closed GPT-3}:
The show \redbox{"Game of Thrones"} was recently green lit for a new
\par
\textbf{Open GPT-3}:
\greenbox{Squid Game}
&
\textbf{June 12, 2022}
\par
\greenbox{Netflix green lights `Squid Game'} season 2...
\par
\textbf{June 17, 2022}
\par
5 things to know for June 17...
\par
\textbf{June 4, 2019}
\par
`Looking for Alaska' details revealed for Hulu limited series...

&
\textbf{February 4, 2022}
\par
The Busch Light Clash goes green this weekend...
\par
\textbf{September 26, 2018}
\par
Dip into 4 new mysteries for fall, including Kate Atkinson's spy novel `Transcription'...
\\

\midrule[.02em]
The IRS announced it will do what this week?
\par
\textbf{Date}: June 24, 2022
\par
\textbf{Answer}: Finish processing the backlog of 2021 tax returns
\par
\textbf{Closed GPT-3}:
The IRS announced it will \redbox{begin processing tax returns} this week.
\par
\textbf{Open GPT-3}:
The IRS announced it will \redbox{begin processing 2021 tax returns} as soon as Jan.\ 24
&
\textbf{January 10, 2022}
\par
\redbox{IRS 2022 tax season set to begin} 2 weeks early on Jan.\ 24... 
\par
\textbf{March 12, 2021}
\par
When will I get my third stimulus check?...
\par
\textbf{March 22, 2021}
\par
IRS says more stimulus checks on the way...

&
\textbf{February 11, 2022}
\par
Don't panic if you got a scary IRS notice...
\par
\textbf{January 10, 2022}
\par
\redbox{IRS will begin processing 2021} \redbox{tax returns} as soon as Jan.\ 24
\\

\midrule[.02em]

Which country is now “bankrupt,” according to a statement this week from its administration?
\par
\textbf{Date}: July 8, 2022
\par
\textbf{Answer}: Sri Lanka
\par
\textbf{Closed GPT-3}:
\redbox{Greece}
\par
\textbf{Open GPT-3}:
\redbox{Venezuela}
&
\textbf{March 2, 2022}
\par
Gun manufacturers are not entirely exempt from being sued... the now-bankrupt gun manufacturer...
\par
\textbf{March 12, 2021}
\par
Mitch McConnell seeks to end Democrat's `crazy policy' of beefed-up unemployment benefits... let states go bankrupt...
&
\textbf{March 20, 2022}
\par
Half of US hotels could close amid coronavirus crisis... hotels around the country go bankrupt...
\par
\textbf{September 26, 2013}
\par
Colo.\ farmers arrested... the now-bankrupt Jensen Farms...
\par
\textbf{January 10, 2022}
\par
Trump administration restrictions on asylum are cruel... Immigration policy is morally bankrupt...
\\

\midrule[.02em]

Which head of state announced his resignation this week?
\par
\textbf{Date}: July 8, 2022
\par
\textbf{Answer}: UK Prime Minister Boris Johnson
\par
\textbf{Closed GPT-3}:
\redbox{Japanese Prime Minister}
\redbox{Shinzo Abe announced his resignation}
\redbox{this week.}
\par
\textbf{Open GPT-3}:
\redbox{Andrew Cuomo}
&
\textbf{August 11, 2021}
\par
NY Gov.\ \redbox{Andrew Cuomo} will resign in two weeks...
\par
\textbf{September 21, 2021}
\par
Maricopa County Supervisor Steve Chucri to resign...
\par
\textbf{January 25, 2016}
\par
Ball State president Ferguson resigns...
&
\textbf{March 23, 2021}
\par
Oregon State University President F.\ King Alexander resigns...
\par
\textbf{August 10, 2021}
\par
NY Gov.\ \redbox{Andrew Cuomo} to resign amid scandal...

\\

\toprule
\end{tabular}
}
\end{center}
\label{tab:examples}
\end{table*}

\section{Related Work}
\label{section:related}
\realtime has time sensitivity, which several prior works addressed on various NLP tasks.
Here we discuss its relation to long-standing summarization and text retrieval tasks, as well as recent work on temporal misalignment between training and evaluation.
We then discuss its connections to dynamic evaluations and open-domain QA.

\noindent\textbf{Summarization/Retrieval over Time} \
\label{section:related_summ_retrieval}
Temporal (or timeline) summarization is a task that retrieves documents from the web and provides their summary \emph{over time} \citep{catizone-etal-2006-evaluating,trec13,trec14,trec15,martschat-markert-2017-improving,martschat-markert-2018-temporally}.
Update summarization \citep{WiKrBe_DUC2007,tac_DangO08} and new event detection/track \citep{allan98,li05_sigir} are tasks that monitor and track newly-added information.
Prior work created datasets and systems for these tasks \citep{TranAN13,tran2015,wang-etal-2015-socially,chen2019ijcai,gholipour-ghalandari-ifrim-2020-examining}.
Their evaluations are usually executed \emph{statically}, with information available at the time of data collection.

In contrast, the TREC real-time summarization track evaluates systems in real time during a 1--2 week evaluation period \citep{trec16,trec17,trec18}.
Several other works and initiatives focused particularly on financial news summarization \citep{filippova-etal-2009-company,passali-etal-2021-towards} or emergency management technology \citep{temnikova-etal-2014-building,ghosh17_smerp,McCreadieBS19}, including the COVID-19 pandemic \citep{BuntainMS20,PasqualiArian2021TANA}.
This work regularly evaluates question answering systems over diverse topics, but we share the goal of dealing with novel and evolving information over time; retrieval or summarization methods from these tasks (e.g., \citealp{yan-etal-2011-timeline,evolutionary_timeline,yan2012,sumblr2013}) can be combined with models in \realtime to serve various time-sensitive information needs from users.
\realtime can also be used to evaluate time-sensitive retrieval systems by the downstream QA performance.

\noindent\textbf{Temporal Misalignment and Degradation} \
\label{section:misalignment}
While not particularly motivated by instantaneous information needs like \realtime, prior work also explored temporal aspects of a variety of NLP tasks.
A flurry of recent work analyzed performance degradation from temporal misalignment between (pre)training and evaluation/deployment on many NLP tasks \citep{LazaridouKGALTG21,rottger-pierrehumbert-2021-temporal-adaptation,waits_for_no_one,onoe2022} and proposed mitigation methods
\citep{huang-paul-2018-examining,huang-paul-2019-neural-temporality,time_aware_lm,temporalwiki,jang2021towards,lee-etal-2022-plug}.
An open-book QA model conditions answer generation upon newly-retrieved documents \citep{rag2020}, but the extent to which answer generation can be updated based on the retrieved documents is limited \citep{longpre-etal-2021-entity}.
Temporal degradation is, therefore, one of the challenges that models in \realtime need to address.

\noindent\textbf{Dynamic Benchmarks} \
\label{section:related_dynamic}
Unlike the majority of datasets in natural language processing, \realtime evaluates systems \textit{dynamically} and its evaluations change over time. 
Several other prior works update challenge test sets \citep{kiela-etal-2021-dynabench,potts-etal-2021-dynasent,dynaboard2021}, evaluation tasks \citep{thrush-etal-2022-dynatask}, or metrics \citep{gem2021,gemv2,MishraA21,billboard}.
\realtime hosts a similar online platform and adopts a dynamic scheme specifically to pursue instantaneous applications.

\noindent\textbf{Open-Domain QA} \
\label{section:related_qa}
Much prior work proposed datasets for open-domain QA for English and beyond \citep{clark-etal-2020-tydi,asai2021xor,asai-etal-2022-mia,longpre2021mkqa,zhang-etal-2021-mr}.
Several recent works challenged the conventional problem setups \citep{chen-yih-2020-open} where correct answers can be found from a fixed, external knowledge source, such as Wikipedia.
Similar to \realtime, \citet{zhang-choi-2021-situatedqa,streamingqa2022} focused on temporal or geographical contexts that can change the answer to the same question.
Consistent with these prior efforts, \realtime aims toward broader applications of question answering beyond the conventional framework.

\section{Conclusion and Future Work}
We introduce \realtime, a dynamic, open-domain QA benchmark that asks questions at the present time.
Our platform announces questions every week and continually evaluates six real-time baselines.
Our experiments from the first year suggest that accurate, up-to-date information retrieval is particularly important to serve speedy information needs.
We hope that \realtime encourages research efforts toward fast, accurate applications of natural language processing.
\section*{Limitations}
This work aims to develop a QA benchmark for addressing instantaneous information needs, including emergency management.
The current version of \realtime has two major limitations due to our annotation framework (\S\ref{section:annotation}): 1) question/answer pairs are all written in English, and the covered topics tend to be English-centric (US and UK); 2) questions are announced on a weekly basis, rather than a truly  instantaneous basis.
Nevertheless, our benchmark departs from many static datasets from prior work and provides an important step towards the research goal.
We hope to develop future versions of \realtime that mitigate these limitations.
\section*{Acknowledgements}
We thank Noriyuki Kojima,  Alisa Liu, Ofir Press, Koji Shiono, Wenya Wang, the ARK group at the UW, and the Mosaic team at the Allen Institute for AI for their helpful feedback on this work.

\bibliographystyle{acl_natbib}
\bibliography{custom}

\begin{thebibliography}{98}
\expandafter\ifx\csname natexlab\endcsname\relax\def\natexlab#1{#1}\fi

\bibitem[{Allan et~al.(2001)Allan, Gupta, and Khandelwal}]{allan01}
James Allan, Rahul Gupta, and Vikas Khandelwal. 2001.
\newblock \href {https://doi.org/10.1145/383952.383954} {Temporal summaries of
  new topics}.
\newblock In \emph{Proc.\ of SIGIR}.

\bibitem[{Allan et~al.(1998)Allan, Papka, and Lavrenko}]{allan98}
James Allan, Ron Papka, and Victor Lavrenko. 1998.
\newblock \href {https://doi.org/10.1145/290941.290954} {On-line new event
  detection and tracking}.
\newblock In \emph{Proc.\ of SIGIR}.

\bibitem[{Alzubi et~al.(2021)Alzubi, Jain, Singh, Parwekar, and
  Gupta}]{Alzubi2021COBERTCQ}
Jafar Ahmad~Abed Alzubi, Rachna Jain, Anubhav Singh, Pritee Parwekar, and Meenu
  Gupta. 2021.
\newblock \href {https://www.ncbi.nlm.nih.gov/pmc/articles/PMC8220121/}
  {{COBERT}: {COVID-19} question answering system using {BERT}}.
\newblock \emph{Arabian Journal for Science and Engineering}.

\bibitem[{Asai et~al.(2021)Asai, Kasai, Clark, Lee, Choi, and
  Hajishirzi}]{asai2021xor}
Akari Asai, Jungo Kasai, Jonathan~H Clark, Kenton Lee, Eunsol Choi, and
  Hannaneh Hajishirzi. 2021.
\newblock \href {https://arxiv.org/abs/2010.11856} {{XOR QA}: Cross-lingual
  open-retrieval question answering}.
\newblock In \emph{Proc.\ of NAACL}.

\bibitem[{Asai et~al.(2022)Asai, Longpre, Kasai, Lee, Zhang, Hu, Yamada, Clark,
  and Choi}]{asai-etal-2022-mia}
Akari Asai, Shayne Longpre, Jungo Kasai, Chia-Hsuan Lee, Rui Zhang, Junjie Hu,
  Ikuya Yamada, Jonathan~H. Clark, and Eunsol Choi. 2022.
\newblock \href {https://aclanthology.org/2022.mia-1.11} {{MIA} 2022 shared
  task: Evaluating cross-lingual open-retrieval question answering for 16
  diverse languages}.
\newblock In \emph{Proc.\ of MIA}.

\bibitem[{Aslam et~al.(2015)Aslam, Diaz, Ekstrand{-}Abueg, McCreadie, Pavlu,
  and Sakai}]{trec15}
Javed~A. Aslam, Fernando Diaz, Matthew Ekstrand{-}Abueg, Richard McCreadie,
  Virgil Pavlu, and Tetsuya Sakai. 2015.
\newblock \href {https://ir.webis.de/anthology/2015.trec_conference-2015.64/}
  {{TREC} 2015 temporal summarization track overview}.
\newblock In \emph{Proc.\ of TREC}.

\bibitem[{Aslam et~al.(2014)Aslam, Ekstrand{-}Abueg, Pavlu, Diaz, McCreadie,
  and Sakai}]{trec14}
Javed~A. Aslam, Matthew Ekstrand{-}Abueg, Virgil Pavlu, Fernando Diaz, Richard
  McCreadie, and Tetsuya Sakai. 2014.
\newblock \href
  {https://trec.nist.gov/pubs/trec23/papers/overview-tempsumm.pdf} {{TREC} 2014
  temporal summarization track overview}.
\newblock In \emph{Proc.\ of TREC}.

\bibitem[{Aslam et~al.(2013)Aslam, Ekstrand{-}Abueg, Pavlu, Diaz, and
  Sakai}]{trec13}
Javed~A. Aslam, Matthew Ekstrand{-}Abueg, Virgil Pavlu, Fernando Diaz, and
  Tetsuya Sakai. 2013.
\newblock \href {http://trec.nist.gov/pubs/trec22/papers/TS.OVERVIEW.pdf}
  {{TREC} 2013 temporal summarization}.
\newblock In \emph{Proc.\ of TREC}.

\bibitem[{Berant et~al.(2013)Berant, Chou, Frostig, and
  Liang}]{berant-etal-2013-semantic}
Jonathan Berant, Andrew Chou, Roy Frostig, and Percy Liang. 2013.
\newblock \href {https://aclanthology.org/D13-1160} {Semantic parsing on
  {F}reebase from question-answer pairs}.
\newblock In \emph{Proc.\ of EMNLP}.

\bibitem[{Brown et~al.(2020)Brown, Mann, Ryder, Subbiah, Kaplan, Dhariwal,
  Neelakantan, Shyam, Sastry, Askell, Agarwal, Herbert-Voss, Krueger, Henighan,
  Child, Ramesh, Ziegler, Wu, Winter, Hesse, Chen, Sigler, Litwin, Gray, Chess,
  Clark, Berner, McCandlish, Radford, Sutskever, and Amodei}]{gpt3}
Tom Brown, Benjamin Mann, Nick Ryder, Melanie Subbiah, Jared~D Kaplan, Prafulla
  Dhariwal, Arvind Neelakantan, Pranav Shyam, Girish Sastry, Amanda Askell,
  Sandhini Agarwal, Ariel Herbert-Voss, Gretchen Krueger, Tom Henighan, Rewon
  Child, Aditya Ramesh, Daniel Ziegler, Jeffrey Wu, Clemens Winter, Chris
  Hesse, Mark Chen, Eric Sigler, Mateusz Litwin, Scott Gray, Benjamin Chess,
  Jack Clark, Christopher Berner, Sam McCandlish, Alec Radford, Ilya Sutskever,
  and Dario Amodei. 2020.
\newblock \href
  {https://proceedings.neurips.cc/paper/2020/file/1457c0d6bfcb4967418bfb8ac142f64a-Paper.pdf}
  {Language models are few-shot learners}.
\newblock In \emph{Proc.\ of NeurIPS}.

\bibitem[{Buntain et~al.(2020)Buntain, McCreadie, and Soboroff}]{BuntainMS20}
Cody Buntain, Richard McCreadie, and Ian Soboroff. 2020.
\newblock \href {https://trec.nist.gov/pubs/trec29/papers/OVERVIEW.IS.pdf}
  {Incident streams 2020: {TRECIS} in the time of {COVID-19}}.
\newblock In \emph{Proc.\ of TREC}.

\bibitem[{Catizone et~al.(2006)Catizone, Dalli, and
  Wilks}]{catizone-etal-2006-evaluating}
Roberta Catizone, Angelo Dalli, and Yorick Wilks. 2006.
\newblock \href {http://www.lrec-conf.org/proceedings/lrec2006/pdf/702_pdf.pdf}
  {Evaluating automatically generated timelines from the web}.
\newblock In \emph{Proc.\ of LREC}.

\bibitem[{Chen et~al.(2017)Chen, Fisch, Weston, and Bordes}]{chen2017reading}
Danqi Chen, Adam Fisch, Jason Weston, and Antoine Bordes. 2017.
\newblock \href {https://www.aclweb.org/anthology/P17-1171} {Reading
  {Wikipedia} to answer open-domain questions}.
\newblock In \emph{Proc.\ of ACL}.

\bibitem[{Chen and Yih(2020)}]{chen-yih-2020-open}
Danqi Chen and Wen-tau Yih. 2020.
\newblock \href {https://www.aclweb.org/anthology/2020.acl-tutorials.8}
  {Open-domain question answering}.
\newblock In \emph{Proc.\ of ACL: Tutorial Abstracts}.

\bibitem[{Chen et~al.(2021)Chen, Wang, and Wang}]{time-sensitive}
Wenhu Chen, Xinyi Wang, and William~Yang Wang. 2021.
\newblock \href {https://arxiv.org/abs/2108.06314} {A dataset for answering
  time-sensitive questions}.
\newblock In \emph{Proc.\ of NeurIPS Datasets and Benchmarks}.

\bibitem[{Chen et~al.(2019)Chen, Chan, Gao, Yu, Zhao, and Yan}]{chen2019ijcai}
Xiuying Chen, Zhangming Chan, Shen Gao, Meng-Hsuan Yu, Dongyan Zhao, and Rui
  Yan. 2019.
\newblock \href {https://doi.org/10.24963/ijcai.2019/686} {Learning towards
  abstractive timeline summarization}.
\newblock In \emph{Proc.\ of IJCAI}.

\bibitem[{Clark et~al.(2020)Clark, Choi, Collins, Garrette, Kwiatkowski,
  Nikolaev, and Palomaki}]{clark-etal-2020-tydi}
Jonathan~H. Clark, Eunsol Choi, Michael Collins, Dan Garrette, Tom Kwiatkowski,
  Vitaly Nikolaev, and Jennimaria Palomaki. 2020.
\newblock \href {https://aclanthology.org/2020.tacl-1.30} {{T}y{D}i {QA}: A
  benchmark for information-seeking question answering in typologically diverse
  languages}.
\newblock \emph{TACL}.

\bibitem[{Dang and Owczarzak(2008)}]{tac_DangO08}
Hoa~Trang Dang and Karolina Owczarzak. 2008.
\newblock \href
  {https://tac.nist.gov/publications/2008/additional.papers/update\_summ\_overview08.proceedings.pdf}
  {Overview of the {TAC} 2008 update summarization task}.
\newblock In \emph{Proc.\ of TAC}.

\bibitem[{Devlin et~al.(2019)Devlin, Chang, Lee, and
  Toutanova}]{devlins2019bert}
Jacob Devlin, Ming-Wei Chang, Kenton Lee, and Kristina Toutanova. 2019.
\newblock \href {https://arxiv.org/abs/810.04805} {{BERT}: Pre-training of deep
  bidirectional transformers for language understanding}.
\newblock In \emph{Proc. of NAACL}.

\bibitem[{Dhingra et~al.(2022)Dhingra, Cole, Eisenschlos, Gillick, Eisenstein,
  and Cohen}]{time_aware_lm}
Bhuwan Dhingra, Jeremy~R. Cole, Julian~Martin Eisenschlos, Daniel Gillick,
  Jacob Eisenstein, and William~W. Cohen. 2022.
\newblock \href {https://doi.org/10.1162/tacl\_a\_00459} {Time-aware language
  models as temporal knowledge bases}.
\newblock \emph{TACL}.

\bibitem[{Evans et~al.(2004)Evans, Klavans, and
  McKeown}]{evans-etal-2004-columbia}
David~Kirk Evans, Judith~L. Klavans, and Kathleen~R. McKeown. 2004.
\newblock \href {https://aclanthology.org/N04-3001/} {{C}olumbia {N}ewsblaster:
  Multilingual news summarization on the web}.
\newblock In \emph{Proc.\ of NAACL: Demonstrations}.

\bibitem[{Filippova et~al.(2009)Filippova, Surdeanu, Ciaramita, and
  Zaragoza}]{filippova-etal-2009-company}
Katja Filippova, Mihai Surdeanu, Massimiliano Ciaramita, and Hugo Zaragoza.
  2009.
\newblock \href {https://aclanthology.org/E09-1029} {Company-oriented
  extractive summarization of financial news}.
\newblock In \emph{Proc.\ of EACL}.

\bibitem[{Gehrmann et~al.(2021)Gehrmann, Adewumi, Aggarwal, Ammanamanchi,
  Anuoluwapo, Bosselut, Chandu, Clinciu, Das, Dhole, Du, Durmus, Dusek, Emezue,
  Gangal, Garbacea, Hashimoto, Hou, Jernite, Jhamtani, Ji, Jolly, Kumar,
  Ladhak, Madaan, Maddela, Mahajan, Mahamood, Majumder, Martins,
  McMillan{-}Major, Mille, van Miltenburg, Nadeem, Narayan, Nikolaev,
  Niyongabo, Osei, Parikh, Perez{-}Beltrachini, Rao, Raunak, Rodriguez,
  Santhanam, Sedoc, Sellam, Shaikh, Shimorina, Cabezudo, Strobelt, Subramani,
  Xu, Yang, Yerukola, and Zhou}]{gem2021}
Sebastian Gehrmann, Tosin~P. Adewumi, Karmanya Aggarwal, Pawan~Sasanka
  Ammanamanchi, Aremu Anuoluwapo, Antoine Bosselut, Khyathi~Raghavi Chandu,
  Miruna{-}Adriana Clinciu, Dipanjan Das, Kaustubh~D. Dhole, Wanyu Du, Esin
  Durmus, Ondrej Dusek, Chris Emezue, Varun Gangal, Cristina Garbacea,
  Tatsunori Hashimoto, Yufang Hou, Yacine Jernite, Harsh Jhamtani, Yangfeng Ji,
  Shailza Jolly, Dhruv Kumar, Faisal Ladhak, Aman Madaan, Mounica Maddela,
  Khyati Mahajan, Saad Mahamood, Bodhisattwa~Prasad Majumder, Pedro~Henrique
  Martins, Angelina McMillan{-}Major, Simon Mille, Emiel van Miltenburg, Moin
  Nadeem, Shashi Narayan, Vitaly Nikolaev, Rubungo~Andre Niyongabo, Salomey
  Osei, Ankur~P. Parikh, Laura Perez{-}Beltrachini, Niranjan~Ramesh Rao, Vikas
  Raunak, Juan~Diego Rodriguez, Sashank Santhanam, Jo{\~{a}}o Sedoc, Thibault
  Sellam, Samira Shaikh, Anastasia Shimorina, Marco Antonio~Sobrevilla
  Cabezudo, Hendrik Strobelt, Nishant Subramani, Wei Xu, Diyi Yang, Akhila
  Yerukola, and Jiawei Zhou. 2021.
\newblock \href {https://arxiv.org/abs/2102.01672} {The {GEM} benchmark:
  Natural language generation, its evaluation and metrics}.
\newblock In \emph{Proc.\ of GEM}.

\bibitem[{Gehrmann et~al.(2022)Gehrmann, Bhattacharjee, Mahendiran, Wang,
  Papangelis, Madaan, McMillan{-}Major, Shvets, Upadhyay, Yao, Wilie,
  Bhagavatula, You, Thomson, Garbacea, Wang, Deutsch, Xiong, Jin, Gkatzia,
  Radev, Clark, Durmus, Ladhak, Ginter, Winata, Strobelt, Hayashi, Novikova,
  Kanerva, Chim, Zhou, Clive, Maynez, Sedoc, Juraska, Dhole, Chandu,
  Perez{-}Beltrachini, Ribeiro, Tunstall, Zhang, Pushkarna, Creutz, White,
  Kale, Eddine, Daheim, Subramani, Dusek, Liang, Ammanamanchi, Zhu, Puduppully,
  Kriz, Shahriyar, Cardenas, Mahamood, Osei, Cahyawijaya, Stajner, Montella,
  Jolly, Mille, Hasan, Shen, AMahidewumi, Raunak, Raheja, Nikolaev, Tsai,
  Jernite, Xu, Sang, Liu, and Hou}]{gemv2}
Sebastian Gehrmann, Abhik Bhattacharjee, Abinaya Mahendiran, Alex Wang,
  Alexandros Papangelis, Aman Madaan, Angelina McMillan{-}Major, Anna Shvets,
  Ashish Upadhyay, Bingsheng Yao, Bryan Wilie, Chandra Bhagavatula, Chaobin
  You, Craig Thomson, Cristina Garbacea, Dakuo Wang, Daniel Deutsch, Deyi
  Xiong, Di~Jin, Dimitra Gkatzia, Dragomir Radev, Elizabeth Clark, Esin Durmus,
  Faisal Ladhak, Filip Ginter, Genta~Indra Winata, Hendrik Strobelt, Hiroaki
  Hayashi, Jekaterina Novikova, Jenna Kanerva, Jenny Chim, Jiawei Zhou, Jordan
  Clive, Joshua Maynez, Jo{\~{a}}o Sedoc, Juraj Juraska, Kaustubh~D. Dhole,
  Khyathi~Raghavi Chandu, Laura Perez{-}Beltrachini, Leonardo F.~R. Ribeiro,
  Lewis Tunstall, Li~Zhang, Mahima Pushkarna, Mathias Creutz, Michael White,
  Mihir~Sanjay Kale, Moussa~Kamal Eddine, Nico Daheim, Nishant Subramani,
  Ondrej Dusek, Paul~Pu Liang, Pawan~Sasanka Ammanamanchi, Qi~Zhu, Ratish
  Puduppully, Reno Kriz, Rifat Shahriyar, Ronald Cardenas, Saad Mahamood,
  Salomey Osei, Samuel Cahyawijaya, Sanja Stajner, S{\'{e}}bastien Montella,
  Shailza Jolly, Simon Mille, Tahmid Hasan, Tianhao Shen, Tosin~P. AMahidewumi,
  Vikas Raunak, Vipul Raheja, Vitaly Nikolaev, Vivian Tsai, Yacine Jernite,
  Ying Xu, Yisi Sang, Yixin Liu, and Yufang Hou. 2022.
\newblock \href {https://arxiv.org/abs/2206.11249} {{GEM}v2: Multilingual {NLG}
  benchmarking in a single line of code}.

\bibitem[{Gholipour~Ghalandari and
  Ifrim(2020)}]{gholipour-ghalandari-ifrim-2020-examining}
Demian Gholipour~Ghalandari and Georgiana Ifrim. 2020.
\newblock \href {https://arxiv.org/abs/2005.10107} {Examining the
  state-of-the-art in news timeline summarization}.
\newblock In \emph{Proc.\ of ACL}.

\bibitem[{Ghosh et~al.(2017)Ghosh, Ghosh, Ganguly, Chakraborty, Jones, and
  Moens}]{ghosh17_smerp}
Saptarshi Ghosh, Kripabandhu Ghosh, Debasis Ganguly, Tanmoy Chakraborty,
  Gareth~J.F. Jones, and Marie-Francine Moens. 2017.
\newblock \href {https://doi.org/10.1145/3130332.3130338} {{ECIR} 2017 workshop
  on exploitation of social media for emergency relief and preparedness
  ({SMERP} 2017)}.
\newblock \emph{SIGIR Forum}.

\bibitem[{Guu et~al.(2020)Guu, Lee, Tung, Pasupat, and Chang}]{guu2020realm}
Kelvin Guu, Kenton Lee, Zora Tung, Panupong Pasupat, and Ming-Wei Chang. 2020.
\newblock \href {https://arxiv.org/abs/2002.08909} {{REALM}:
  Retrieval-augmented language model pre-training}.
\newblock In \emph{Proc.\ of ICML}.

\bibitem[{Hermann et~al.(2015)Hermann, Kocisky, Grefenstette, Espeholt, Kay,
  Suleyman, and Blunsom}]{teaching_machines}
Karl~Moritz Hermann, Tomas Kocisky, Edward Grefenstette, Lasse Espeholt, Will
  Kay, Mustafa Suleyman, and Phil Blunsom. 2015.
\newblock \href {https://arxiv.org/abs/1506.03340} {Teaching machines to read
  and comprehend}.
\newblock In \emph{Proc. of NeurIPS}.

\bibitem[{Huang and Paul(2018)}]{huang-paul-2018-examining}
Xiaolei Huang and Michael~J. Paul. 2018.
\newblock \href {https://aclanthology.org/P18-2110} {Examining temporality in
  document classification}.
\newblock In \emph{Proc.\ of ACL}.

\bibitem[{Huang and Paul(2019)}]{huang-paul-2019-neural-temporality}
Xiaolei Huang and Michael~J. Paul. 2019.
\newblock \href {https://aclanthology.org/P19-1403} {Neural temporality
  adaptation for document classification: Diachronic word embeddings and domain
  adaptation models}.
\newblock In \emph{Proc.\ of ACL}.

\bibitem[{Imran et~al.(2015)Imran, Castillo, Diaz, and Vieweg}]{imran2015}
Muhammad Imran, Carlos Castillo, Fernando Diaz, and Sarah Vieweg. 2015.
\newblock \href {https://doi.org/10.1145/2771588} {Processing social media
  messages in mass emergency: A survey}.
\newblock \emph{ACM Computing Surveys}.

\bibitem[{Imran et~al.(2013)Imran, Elbassuoni, Castillo, Diaz, and
  Meier}]{imran2013practical}
Muhammad Imran, Shady Elbassuoni, Carlos Castillo, Fernando Diaz, and Patrick
  Meier. 2013.
\newblock \href {https://dl.acm.org/doi/10.1145/2487788.2488109} {Practical
  extraction of disaster-relevant information from social media}.
\newblock In \emph{Proc.\ of WWW}.

\bibitem[{Imran et~al.(2016)Imran, Mitra, and Castillo}]{imran2016lrec}
Muhammad Imran, Prasenjit Mitra, and Carlos Castillo. 2016.
\newblock \href {https://arxiv.org/abs/1605.05894} {Twitter as a lifeline:
  Human-annotated {Twitte}r corpora for {NLP} of crisis-related messages}.
\newblock In \emph{Proc.\ of LREC}.

\bibitem[{Izacard and Grave(2021)}]{izacard2021leveraging}
Gautier Izacard and Edouard Grave. 2021.
\newblock \href {https://arxiv.org/abs/2007.01282} {Leveraging passage
  retrieval with generative models for open domain question answering}.
\newblock In \emph{Proc. of EACL}.

\bibitem[{Jang et~al.(2022{\natexlab{a}})Jang, Ye, Lee, Yang, Shin, Han, Kim,
  and Seo}]{temporalwiki}
Joel Jang, Seonghyeon Ye, Changho Lee, Sohee Yang, Joongbo Shin, Janghoon Han,
  Gyeonghun Kim, and Minjoon Seo. 2022{\natexlab{a}}.
\newblock \href {https://arxiv.org/abs/2204.14211} {{TemporalWiki}: {A}
  lifelong benchmark for training and evaluating ever-evolving language
  models}.

\bibitem[{Jang et~al.(2022{\natexlab{b}})Jang, Ye, Yang, Shin, Han, Kim, Choi,
  and Seo}]{jang2021towards}
Joel Jang, Seonghyeon Ye, Sohee Yang, Joongbo Shin, Janghoon Han, Gyeonghun
  Kim, Stanley~Jungkyu Choi, and Minjoon Seo. 2022{\natexlab{b}}.
\newblock \href {https://arxiv.org/abs/2110.03215} {Towards continual knowledge
  learning of language models}.
\newblock In \emph{Proc.\ of ICLR}.

\bibitem[{Jia et~al.(2018)Jia, Abujabal, Roy, Str{\"{o}}tgen, and
  Weikum}]{tempquestions18}
Zhen Jia, Abdalghani Abujabal, Rishiraj~Saha Roy, Jannik Str{\"{o}}tgen, and
  Gerhard Weikum. 2018.
\newblock \href {https://doi.org/10.1145/3184558.3191536} {{TempQuestions}: {A}
  benchmark for temporal question answering}.
\newblock In \emph{Companion of the WWW}.

\bibitem[{Joshi et~al.(2017)Joshi, Choi, Weld, and
  Zettlemoyer}]{joshi-etal-2017-triviaqa}
Mandar Joshi, Eunsol Choi, Daniel Weld, and Luke Zettlemoyer. 2017.
\newblock \href {https://arxiv.org/abs/1705.03551} {{T}rivia{QA}: A large scale
  distantly supervised challenge dataset for reading comprehension}.
\newblock In \emph{Proc.\ of ACL}.

\bibitem[{Karpukhin et~al.(2020)Karpukhin, Oguz, Min, Lewis, Wu, Edunov, Chen,
  and Yih}]{karpukhin-etal-2020-dense}
Vladimir Karpukhin, Barlas Oguz, Sewon Min, Patrick Lewis, Ledell Wu, Sergey
  Edunov, Danqi Chen, and Wen-tau Yih. 2020.
\newblock \href {https://arxiv.org/abs/2004.04906} {Dense passage retrieval for
  open-domain question answering}.
\newblock In \emph{Proc.\ of EMNLP}.

\bibitem[{Kasai et~al.(2022)Kasai, Sakaguchi, Bras, Dunagan, Morrison, Fabbri,
  Choi, and Smith}]{billboard}
Jungo Kasai, Keisuke Sakaguchi, Ronan~Le Bras, Lavinia Dunagan, Jacob Morrison,
  Alexander~R. Fabbri, Yejin Choi, and Noah~A. Smith. 2022.
\newblock \href {https://arxiv.org/abs/2112.04139} {Bidimensional leaderboards:
  Generate and evaluate language hand in hand}.
\newblock In \emph{Proc.\ of NAACL}.

\bibitem[{Kiela et~al.(2021)Kiela, Bartolo, Nie, Kaushik, Geiger, Wu, Vidgen,
  Prasad, Singh, Ringshia, Ma, Thrush, Riedel, Waseem, Stenetorp, Jia, Bansal,
  Potts, and Williams}]{kiela-etal-2021-dynabench}
Douwe Kiela, Max Bartolo, Yixin Nie, Divyansh Kaushik, Atticus Geiger,
  Zhengxuan Wu, Bertie Vidgen, Grusha Prasad, Amanpreet Singh, Pratik Ringshia,
  Zhiyi Ma, Tristan Thrush, Sebastian Riedel, Zeerak Waseem, Pontus Stenetorp,
  Robin Jia, Mohit Bansal, Christopher Potts, and Adina Williams. 2021.
\newblock \href {https://arxiv.org/abs/2104.14337} {Dynabench: Rethinking
  benchmarking in {NLP}}.
\newblock In \emph{Proc.\ of NAACL}.

\bibitem[{Kwiatkowski et~al.(2019)Kwiatkowski, Palomaki, Redfield, Collins,
  Parikh, Alberti, Epstein, Polosukhin, Devlin, Lee
  et~al.}]{kwiatkowski2019natural}
Tom Kwiatkowski, Jennimaria Palomaki, Olivia Redfield, Michael Collins, Ankur
  Parikh, Chris Alberti, Danielle Epstein, Illia Polosukhin, Jacob Devlin,
  Kenton Lee, et~al. 2019.
\newblock \href {https://research.google/pubs/pub47761/} {Natural questions: a
  benchmark for question answering research}.
\newblock \emph{TACL}.

\bibitem[{Lai et~al.(2017)Lai, Xie, Liu, Yang, and Hovy}]{lai-etal-2017-race}
Guokun Lai, Qizhe Xie, Hanxiao Liu, Yiming Yang, and Eduard Hovy. 2017.
\newblock \href {https://arxiv.org/abs/1704.04683} {{RACE}: Large-scale
  {R}e{A}ding comprehension dataset from examinations}.
\newblock In \emph{Proc.\ of EMNLP}.

\bibitem[{Lazaridou et~al.(2022)Lazaridou, Gribovskaya, Stokowiec, and
  Grigorev}]{internet_gpt3}
Angeliki Lazaridou, Elena Gribovskaya, Wojciech Stokowiec, and Nikolai
  Grigorev. 2022.
\newblock \href {https://arxiv.org/abs/2203.05115} {Internet-augmented language
  models through few-shot prompting for open-domain question answering}.

\bibitem[{Lazaridou et~al.(2021)Lazaridou, Kuncoro, Gribovskaya, Agrawal,
  Liska, Terzi, Gimenez, de~Masson~d'Autume, Kocisk{\'{y}}, Ruder, Yogatama,
  Cao, Young, and Blunsom}]{LazaridouKGALTG21}
Angeliki Lazaridou, Adhiguna Kuncoro, Elena Gribovskaya, Devang Agrawal, Adam
  Liska, Tayfun Terzi, Mai Gimenez, Cyprien de~Masson~d'Autume, Tom{\'{a}}s
  Kocisk{\'{y}}, Sebastian Ruder, Dani Yogatama, Kris Cao, Susannah Young, and
  Phil Blunsom. 2021.
\newblock \href {https://arxiv.org/abs/2102.01951} {Mind the gap: Assessing
  temporal generalization in neural language models}.
\newblock In \emph{Proc.\ of NeurIPS}.

\bibitem[{Lee et~al.(2020)Lee, Yi, Jeong, Sung, Yoon, Choi, Ko, and
  Kang}]{lee-etal-2020-answering}
Jinhyuk Lee, Sean~S. Yi, Minbyul Jeong, Mujeen Sung, WonJin Yoon, Yonghwa Choi,
  Miyoung Ko, and Jaewoo Kang. 2020.
\newblock \href {https://aclanthology.org/2020.nlpcovid19-2.1} {Answering
  questions on {COVID}-19 in real-time}.
\newblock In \emph{Proc.\ of the 1st Workshop on {NLP} for {COVID}-19 (Part 2)
  at {EMNLP} 2020}.

\bibitem[{Lee et~al.(2022)Lee, Han, Hwang, Lee, Park, and
  Lee}]{lee-etal-2022-plug}
Kyungjae Lee, Wookje Han, Seung-won Hwang, Hwaran Lee, Joonsuk Park, and
  Sang-Woo Lee. 2022.
\newblock \href {https://aclanthology.org/2022.findings-acl.37/} {Plug-and-play
  adaptation for continuously-updated {QA}}.
\newblock In \emph{Findings of the ACL: ACL 2022}.

\bibitem[{Lewis et~al.(2020{\natexlab{a}})Lewis, Liu, Goyal, Ghazvininejad,
  Mohamed, Levy, Stoyanov, and Zettlemoyer}]{lewis-etal-2020-bart}
Mike Lewis, Yinhan Liu, Naman Goyal, Marjan Ghazvininejad, Abdelrahman Mohamed,
  Omer Levy, Veselin Stoyanov, and Luke Zettlemoyer. 2020{\natexlab{a}}.
\newblock \href {https://www.aclweb.org/anthology/2020.acl-main.703} {{BART}:
  Denoising sequence-to-sequence pre-training for natural language generation,
  translation, and comprehension}.
\newblock In \emph{Proc.\ of ACL}.

\bibitem[{Lewis et~al.(2020{\natexlab{b}})Lewis, Perez, Piktus, Petroni,
  Karpukhin, Goyal, K{\"{u}}ttler, Lewis, Yih, Rockt{\"{a}}schel, Riedel, and
  Kiela}]{rag2020}
Patrick S.~H. Lewis, Ethan Perez, Aleksandra Piktus, Fabio Petroni, Vladimir
  Karpukhin, Naman Goyal, Heinrich K{\"{u}}ttler, Mike Lewis, Wen{-}tau Yih,
  Tim Rockt{\"{a}}schel, Sebastian Riedel, and Douwe Kiela. 2020{\natexlab{b}}.
\newblock \href {https://arxiv.org/abs/2005.11401} {Retrieval-augmented
  generation for knowledge-intensive {NLP} tasks}.
\newblock In \emph{Proc.\ of NeurIPS}.

\bibitem[{Li et~al.(2005)Li, Wang, Li, and Ma}]{li05_sigir}
Zhiwei Li, Bin Wang, Mingjing Li, and Wei-Ying Ma. 2005.
\newblock \href {https://doi.org/10.1145/1076034.1076055} {A probabilistic
  model for retrospective news event detection}.
\newblock In \emph{Proc.\ of SIGIR}.

\bibitem[{Lin et~al.(2017)Lin, Mohammed, Sequiera, Tan, Ghelani, Abualsaud,
  McCreadie, Milajevs, and Voorhees}]{trec17}
Jimmy Lin, Salman Mohammed, Royal Sequiera, Luchen Tan, Nimesh Ghelani, Mustafa
  Abualsaud, Richard McCreadie, Dmitrijs Milajevs, and Ellen~M. Voorhees. 2017.
\newblock \href {https://trec.nist.gov/pubs/trec26/papers/Overview-RT.pdf}
  {Overview of the {TREC} 2017 real-time summarization track}.
\newblock In \emph{Proc.\ of TREC}.

\bibitem[{Lin et~al.(2016)Lin, Roegiest, Tan, McCreadie, Voorhees, and
  Diaz}]{trec16}
Jimmy Lin, Adam Roegiest, Luchen Tan, Richard McCreadie, Ellen~M. Voorhees, and
  Fernando Diaz. 2016.
\newblock \href
  {https://cs.uwaterloo.ca/~jimmylin/publications/Lin_etal_TREC2016.pdf}
  {Overview of the {TREC} 2016 real-time summarization track}.
\newblock In \emph{Proc.\ of TREC}.

\bibitem[{Li{\v{s}}ka et~al.(2022)Li{\v{s}}ka, Ko{\v{c}}isk{\'y}, Gribovskaya,
  Terzi, Sezener, Agrawal, de~Masson~d'Autume, Scholtes, Zaheer, Young, Austin,
  Blunsom, and Lazaridou}]{streamingqa2022}
Adam Li{\v{s}}ka, Tom{\'a}{\v{s}} Ko{\v{c}}isk{\'y}, Elena Gribovskaya, Tayfun
  Terzi, Eren Sezener, Devang Agrawal, Cyprien de~Masson~d'Autume, Tim
  Scholtes, Manzil Zaheer, Susannah Young, Ellen Gilsenan-McMahon~Sophia
  Austin, Phil Blunsom, and Angeliki Lazaridou. 2022.
\newblock \href {https://arxiv.org/abs/2205.11388} {{StreamingQA}: A benchmark
  for adaptation to new knowledge over time in question answering models}.

\bibitem[{Longpre et~al.(2021{\natexlab{a}})Longpre, Lu, and
  Daiber}]{longpre2021mkqa}
Shayne Longpre, Yi~Lu, and Joachim Daiber. 2021{\natexlab{a}}.
\newblock \href {https://arxiv.org/abs/2007.15207} {{MKQA}: A linguistically
  diverse benchmark for multilingual open domain question answering}.
\newblock \emph{TACL}.

\bibitem[{Longpre et~al.(2021{\natexlab{b}})Longpre, Perisetla, Chen, Ramesh,
  DuBois, and Singh}]{longpre-etal-2021-entity}
Shayne Longpre, Kartik Perisetla, Anthony Chen, Nikhil Ramesh, Chris DuBois,
  and Sameer Singh. 2021{\natexlab{b}}.
\newblock \href {https://arxiv.org/abs/2109.05052} {Entity-based knowledge
  conflicts in question answering}.
\newblock In \emph{Proc.\ of EMNLP}.

\bibitem[{Luu et~al.(2022)Luu, Khashabi, Gururangan, Mandyam, and
  Smith}]{waits_for_no_one}
Kelvin Luu, Daniel Khashabi, Suchin Gururangan, Karishma Mandyam, and Noah~A.
  Smith. 2022.
\newblock \href {https://arxiv.org/abs/2111.07408} {Time waits for no one!
  analysis and challenges of temporal misalignment}.
\newblock In \emph{Proc.\ of NAACL}.

\bibitem[{Ma et~al.(2021)Ma, Ethayarajh, Thrush, Jain, Wu, Jia, Potts,
  Williams, and Kiela}]{dynaboard2021}
Zhiyi Ma, Kawin Ethayarajh, Tristan Thrush, Somya Jain, Ledell Wu, Robin Jia,
  Christopher Potts, Adina Williams, and Douwe Kiela. 2021.
\newblock \href {https://arxiv.org/abs/2106.06052} {Dynaboard: An
  evaluation-as-a-service platform for holistic next-generation benchmarking}.
\newblock In \emph{Proc.\ of NeurIPS}.

\bibitem[{Martschat and Markert(2017)}]{martschat-markert-2017-improving}
Sebastian Martschat and Katja Markert. 2017.
\newblock \href {https://aclanthology.org/E17-2046/} {Improving {ROUGE} for
  timeline summarization}.
\newblock In \emph{Proc.\ of EACL}.

\bibitem[{Martschat and Markert(2018)}]{martschat-markert-2018-temporally}
Sebastian Martschat and Katja Markert. 2018.
\newblock \href {https://arxiv.org/abs/1810.07949} {A temporally sensitive
  submodularity framework for timeline summarization}.
\newblock In \emph{Proc.\ of CoNLL}.

\bibitem[{McCreadie et~al.(2019)McCreadie, Buntain, and
  Soboroff}]{McCreadieBS19}
Richard McCreadie, Cody Buntain, and Ian Soboroff. 2019.
\newblock \href
  {http://idl.iscram.org/files/richardmccreadie/2019/1867\_RichardMcCreadie\_etal2019.pdf}
  {{TREC} incident streams: Finding actionable information on social media}.
\newblock In \emph{Proc.\ of ISCRAM}.

\bibitem[{McKeown et~al.(2003)McKeown, Barzilay, Chen, Elson, Evans, Klavans,
  Nenkova, Schiffman, and Sigelman}]{mckeown-etal-2003-columbias}
Kathleen McKeown, Regina Barzilay, John Chen, David Elson, David Evans, Judith
  Klavans, Ani Nenkova, Barry Schiffman, and Sergey Sigelman. 2003.
\newblock \href {https://aclanthology.org/N03-4008} {{C}olumbia{'}s
  newsblaster: New features and future directions}.
\newblock In \emph{Proc.\ of NAACL: Demonstrations}.

\bibitem[{McKeown et~al.(2002)McKeown, Barzilay, Evans, Hatzivassiloglou,
  Klavans, Nenkova, Sable, Schiffman, and Sigelman}]{tracking_newsblaster}
Kathleen~R. McKeown, Regina Barzilay, David Evans, Vasileios Hatzivassiloglou,
  Judith~L. Klavans, Ani Nenkova, Carl Sable, Barry Schiffman, and Sergey
  Sigelman. 2002.
\newblock \href {https://dl.acm.org/doi/10.5555/1289189.1289212} {Tracking and
  summarizing news on a daily basis with {Columbia's Newsblaster}}.
\newblock In \emph{Proc.\ of HLT}.

\bibitem[{Min et~al.(2019)Min, Chen, Zettlemoyer, and
  Hajishirzi}]{min2019knowledge}
Sewon Min, Danqi Chen, Luke Zettlemoyer, and Hannaneh Hajishirzi. 2019.
\newblock \href {https://arxiv.org/abs/1911.03868} {Knowledge guided text
  retrieval and reading for open domain question answering}.

\bibitem[{Mishra and Arunkumar(2021)}]{MishraA21}
Swaroop Mishra and Anjana Arunkumar. 2021.
\newblock \href {https://arxiv.org/abs/2106.05532} {How robust are model
  rankings : {A} leaderboard customization approach for equitable evaluation}.
\newblock In \emph{Proc.\ of AAAI}.

\bibitem[{M{\"o}ller et~al.(2020)M{\"o}ller, Reina, Jayakumar, and
  Pietsch}]{moller-etal-2020-covid}
Timo M{\"o}ller, Anthony Reina, Raghavan Jayakumar, and Malte Pietsch. 2020.
\newblock \href {https://aclanthology.org/2020.nlpcovid19-acl.18} {{COVID-QA}:
  A question answering dataset for {COVID}-19}.
\newblock In \emph{Proc.\ of the 1st Workshop on {NLP} for {COVID-19} at {ACL}
  2020}.

\bibitem[{Nguyen et~al.(2016)Nguyen, Al{-}Mannai, Joty, Sajjad, Imran, and
  Mitra}]{nguyen_rapid}
Dat~Tien Nguyen, Kamla Al{-}Mannai, Shafiq~R. Joty, Hassan Sajjad, Muhammad
  Imran, and Prasenjit Mitra. 2016.
\newblock \href {http://arxiv.org/abs/1608.03902} {Rapid classification of
  crisis-related data on social networks using convolutional neural networks}.
\newblock In \emph{Proc.\ of ICWSM}.

\bibitem[{Onoe et~al.(2022)Onoe, Zhang, Choi, and Durrett}]{onoe2022}
Yasumasa Onoe, Michael J.~Q. Zhang, Eunsol Choi, and Greg Durrett. 2022.
\newblock \href {https://arxiv.org/abs/2205.02832} {Entity cloze by date: What
  {LM}s know about unseen entities}.
\newblock In \emph{Findings of the ACL: NAACL 2022}.

\bibitem[{Pasquali et~al.(2021)Pasquali, Campos, Ribeiro, Santana, Jorge, and
  Jatowt}]{PasqualiArian2021TANA}
Arian Pasquali, Ricardo Campos, Alexandre Ribeiro, Brenda Santana, Alípio
  Jorge, and Adam Jatowt. 2021.
\newblock \href
  {https://link.springer.com/chapter/10.1007/978-3-030-72113-8_33}
  {{TLS-Covid19}: A new annotated corpus for timeline summarization}.
\newblock In \emph{Advances in Information Retrieval}.

\bibitem[{Passali et~al.(2021)Passali, Gidiotis, Chatzikyriakidis, and
  Tsoumakas}]{passali-etal-2021-towards}
Tatiana Passali, Alexios Gidiotis, Efstathios Chatzikyriakidis, and Grigorios
  Tsoumakas. 2021.
\newblock \href {https://www.aclweb.org/anthology/2021.hcinlp-1.4} {Towards
  human-centered summarization: A case study on financial news}.
\newblock In \emph{Proc.\ of the First Workshop on Bridging Human{--}Computer
  Interaction and Natural Language Processing}.

\bibitem[{Potts et~al.(2021)Potts, Wu, Geiger, and
  Kiela}]{potts-etal-2021-dynasent}
Christopher Potts, Zhengxuan Wu, Atticus Geiger, and Douwe Kiela. 2021.
\newblock \href {https://arxiv.org/abs/2012.15349} {{D}yna{S}ent: A dynamic
  benchmark for sentiment analysis}.
\newblock In \emph{Proc.\ of ACL}.

\bibitem[{Radev et~al.(2001)Radev, Blair-Goldensohn, Zhang, and
  Raghavan}]{newsisessence}
Dragomir~R. Radev, Sasha Blair-Goldensohn, Zhu Zhang, and Revathi~Sundara
  Raghavan. 2001.
\newblock \href {https://aclanthology.org/H01-1056/} {{NewsInEssence}: A system
  for domain-independent, real-time news clustering and multi-document
  summarization}.
\newblock In \emph{Proc.\ of HLT}.

\bibitem[{Raffel et~al.(2020)Raffel, Shazeer, Roberts, Lee, Narang, Matena,
  Zhou, Li, and Liu}]{2020t5}
Colin Raffel, Noam Shazeer, Adam Roberts, Katherine Lee, Sharan Narang, Michael
  Matena, Yanqi Zhou, Wei Li, and Peter~J. Liu. 2020.
\newblock \href {http://jmlr.org/papers/v21/20-074.html} {Exploring the limits
  of transfer learning with a unified text-to-text transformer}.
\newblock \emph{JMLR}.

\bibitem[{Rajpurkar et~al.(2018)Rajpurkar, Jia, and
  Liang}]{rajpurkar-etal-2018-know}
Pranav Rajpurkar, Robin Jia, and Percy Liang. 2018.
\newblock \href {https://arxiv.org/abs/1806.03822} {Know what you don{'}t know:
  Unanswerable questions for {SQ}u{AD}}.
\newblock In \emph{Proc.\ of ACL}.

\bibitem[{Rajpurkar et~al.(2016)Rajpurkar, Zhang, Lopyrev, and
  Liang}]{rajpurkar-etal-2016-squad}
Pranav Rajpurkar, Jian Zhang, Konstantin Lopyrev, and Percy Liang. 2016.
\newblock \href {https://arxiv.org/abs/1606.05250} {{SQ}u{AD}: 100,000+
  questions for machine comprehension of text}.
\newblock In \emph{Proc.\ of EMNLP}.

\bibitem[{Richardson et~al.(2013)Richardson, Burges, and
  Renshaw}]{richardson-etal-2013-mctest}
Matthew Richardson, Christopher~J.C. Burges, and Erin Renshaw. 2013.
\newblock \href {https://aclanthology.org/D13-1020} {{MCT}est: A challenge
  dataset for the open-domain machine comprehension of text}.
\newblock In \emph{Proc.\ of EMNLP}.

\bibitem[{Roberts et~al.(2020)Roberts, Raffel, and
  Shazeer}]{roberts-etal-2020-much}
Adam Roberts, Colin Raffel, and Noam Shazeer. 2020.
\newblock \href {https://arxiv.org/abs/2002.08910} {How much knowledge can you
  pack into the parameters of a language model?}
\newblock In \emph{Proc.\ of EMNLP}.

\bibitem[{R{\"o}ttger and
  Pierrehumbert(2021)}]{rottger-pierrehumbert-2021-temporal-adaptation}
Paul R{\"o}ttger and Janet Pierrehumbert. 2021.
\newblock \href {https://arxiv.org/abs/2104.08116} {Temporal adaptation of
  {BERT} and performance on downstream document classification: Insights from
  social media}.
\newblock In \emph{Findings of the ACL: EMNLP 2021}.

\bibitem[{Sakaguchi et~al.(2020)Sakaguchi, Bras, Bhagavatula, and
  Choi}]{winogrande}
Keisuke Sakaguchi, Ronan~Le Bras, Chandra Bhagavatula, and Yejin Choi. 2020.
\newblock \href {https://arxiv.org/abs/1907.10641} {{WinoGrande}: An
  adversarial {W}inograd schema challenge at scale}.
\newblock In \emph{Proc.\ of AAAI}.

\bibitem[{Seo et~al.(2019)Seo, Lee, Kwiatkowski, Parikh, Farhadi, and
  Hajishirzi}]{seo-etal-2019-real}
Minjoon Seo, Jinhyuk Lee, Tom Kwiatkowski, Ankur Parikh, Ali Farhadi, and
  Hannaneh Hajishirzi. 2019.
\newblock \href {https://arxiv.org/abs/1906.05807} {Real-time open-domain
  question answering with dense-sparse phrase index}.
\newblock In \emph{Proc.\ of ACL}.

\bibitem[{Sequiera et~al.(2018)Sequiera, Tan, and Lin}]{trec18}
Royal Sequiera, Luchen Tan, and Jimmy Lin. 2018.
\newblock \href {https://trec.nist.gov/pubs/trec27/papers/Overview-RTS.pdf}
  {Overview of the {TREC} 2018 real-time summarization track}.
\newblock In \emph{Proc.\ of TREC}.

\bibitem[{Shou et~al.(2013)Shou, Wang, Chen, and Chen}]{sumblr2013}
Lidan Shou, Zhenhua Wang, Ke~Chen, and Gang Chen. 2013.
\newblock \href {https://doi.org/10.1145/2484028.2484045} {Sumblr: Continuous
  summarization of evolving tweet streams}.
\newblock In \emph{Proc.\ of SIGIR}.

\bibitem[{Talmor et~al.(2019)Talmor, Herzig, Lourie, and
  Berant}]{talmor-etal-2019-commonsenseqa}
Alon Talmor, Jonathan Herzig, Nicholas Lourie, and Jonathan Berant. 2019.
\newblock \href {https://arxiv.org/abs/1811.00937} {{C}ommonsense{QA}: A
  question answering challenge targeting commonsense knowledge}.
\newblock In \emph{Proc.\ of NAACL}.

\bibitem[{Temnikova et~al.(2014)Temnikova, Varga, and
  Biyikli}]{temnikova-etal-2014-building}
Irina Temnikova, Andrea Varga, and Dogan Biyikli. 2014.
\newblock \href {https://aclanthology.org/L14-1503/} {Building a crisis
  management term resource for social media: The case of floods and protests}.
\newblock In \emph{Proc.\ of LREC}.

\bibitem[{Thrush et~al.(2022)Thrush, Tirumala, Gupta, Bartolo, Rodriguez, Kane,
  Gaviria~Rojas, Mattson, Williams, and Kiela}]{thrush-etal-2022-dynatask}
Tristan Thrush, Kushal Tirumala, Anmol Gupta, Max Bartolo, Pedro Rodriguez,
  Tariq Kane, William Gaviria~Rojas, Peter Mattson, Adina Williams, and Douwe
  Kiela. 2022.
\newblock \href {https://arxiv.org/abs/2204.01906} {Dynatask: A framework for
  creating dynamic {AI} benchmark tasks}.
\newblock In \emph{Proc.\ of ACL: System Demonstrations}.

\bibitem[{Tran et~al.(2015)Tran, Alrifai, and Herder}]{tran2015}
Giang Tran, Mohammad Alrifai, and Eelco Herder. 2015.
\newblock \href
  {https://www.eelcoherder.com/images/publications/2015/ecir2015_timeline_summarization.pdf}
  {Timeline summarization from relevant headlines}.
\newblock In \emph{Advances in Information Retrieval}.

\bibitem[{Tran et~al.(2013)Tran, Alrifai, and Nguyen}]{TranAN13}
Giang~Binh Tran, Mohammad Alrifai, and Dat~Quoc Nguyen. 2013.
\newblock \href {https://dl.acm.org/doi/10.1145/2487788.2487829} {Predicting
  relevant news events for timeline summaries}.
\newblock In \emph{Proc.\ of WWW Companion}.

\bibitem[{Wang et~al.(2015)Wang, Cardie, and
  Marchetti}]{wang-etal-2015-socially}
Lu~Wang, Claire Cardie, and Galen Marchetti. 2015.
\newblock \href {https://arxiv.org/abs/1606.05699} {Socially-informed timeline
  generation for complex events}.
\newblock In \emph{Proc.\ of NAACL}.

\bibitem[{Wang et~al.(2020)Wang, Lo, Chandrasekhar, Reas, Yang, Burdick, Eide,
  Funk, Katsis, Kinney, Li, Liu, Merrill, Mooney, Murdick, Rishi, Sheehan,
  Shen, Stilson, Wade, Wang, Wang, Wilhelm, Xie, Raymond, Weld, Etzioni, and
  Kohlmeier}]{wang-etal-2020-cord}
Lucy~Lu Wang, Kyle Lo, Yoganand Chandrasekhar, Russell Reas, Jiangjiang Yang,
  Doug Burdick, Darrin Eide, Kathryn Funk, Yannis Katsis, Rodney~Michael
  Kinney, Yunyao Li, Ziyang Liu, William Merrill, Paul Mooney, Dewey~A.
  Murdick, Devvret Rishi, Jerry Sheehan, Zhihong Shen, Brandon Stilson, Alex~D.
  Wade, Kuansan Wang, Nancy Xin~Ru Wang, Christopher Wilhelm, Boya Xie,
  Douglas~M. Raymond, Daniel~S. Weld, Oren Etzioni, and Sebastian Kohlmeier.
  2020.
\newblock \href {https://arxiv.org/abs/2004.10706} {{CORD-19}: The {COVID-19}
  open research dataset}.
\newblock In \emph{Proc.\ of the 1st Workshop on {NLP} for {COVID-19} at {ACL}
  2020}.

\bibitem[{Wang et~al.(2019)Wang, Ng, Ma, Nallapati, and
  Xiang}]{wang-etal-2019-multi}
Zhiguo Wang, Patrick Ng, Xiaofei Ma, Ramesh Nallapati, and Bing Xiang. 2019.
\newblock \href {https://arxiv.org/abs/1908.08167} {Multi-passage {BERT}: A
  globally normalized {BERT} model for open-domain question answering}.
\newblock In \emph{Proc.\ of EMNLP}.

\bibitem[{Witte et~al.(2007)Witte, Krestel, and Bergler}]{WiKrBe_DUC2007}
Ren{\'e} Witte, Ralf Krestel, and Sabine Bergler. 2007.
\newblock \href {http://duc.nist.gov/pubs/2007papers/ukarlsruhe.final.pdf}
  {Generating update summaries for {DUC} 2007}.
\newblock In \emph{Proc.\ of DUC}.

\bibitem[{Wolf et~al.(2020)Wolf, Debut, Sanh, Chaumond, Delangue, Moi, Cistac,
  Rault, Louf, Funtowicz, Davison, Shleifer, von Platen, Ma, Jernite, Plu, Xu,
  Le~Scao, Gugger, Drame, Lhoest, and Rush}]{wolf-etal-2020-transformers}
Thomas Wolf, Lysandre Debut, Victor Sanh, Julien Chaumond, Clement Delangue,
  Anthony Moi, Pierric Cistac, Tim Rault, Remi Louf, Morgan Funtowicz, Joe
  Davison, Sam Shleifer, Patrick von Platen, Clara Ma, Yacine Jernite, Julien
  Plu, Canwen Xu, Teven Le~Scao, Sylvain Gugger, Mariama Drame, Quentin Lhoest,
  and Alexander Rush. 2020.
\newblock \href {https://arxiv.org/abs/1910.03771} {Transformers:
  State-of-the-art natural language processing}.
\newblock In \emph{Proc.\ of EMNLP: System Demonstrations}.

\bibitem[{Yan et~al.(2011{\natexlab{a}})Yan, Kong, Huang, Wan, Li, and
  Zhang}]{yan-etal-2011-timeline}
Rui Yan, Liang Kong, Congrui Huang, Xiaojun Wan, Xiaoming Li, and Yan Zhang.
  2011{\natexlab{a}}.
\newblock \href {https://aclanthology.org/D11-1040} {Timeline generation
  through evolutionary trans-temporal summarization}.
\newblock In \emph{Proc.\ of EMNLP}.

\bibitem[{Yan et~al.(2012)Yan, Wan, Lapata, Zhao, Cheng, and Li}]{yan2012}
Rui Yan, Xiaojun Wan, Mirella Lapata, Wayne~Xin Zhao, Pu-Jen Cheng, and
  Xiaoming Li. 2012.
\newblock \href {https://doi.org/10.1145/2396761.2396799} {Visualizing
  timelines: Evolutionary summarization via iterative reinforcement between
  text and image streams}.
\newblock In \emph{Proc.\ of CIKM}.

\bibitem[{Yan et~al.(2011{\natexlab{b}})Yan, Wan, Otterbacher, Kong, Li, and
  Zhang}]{evolutionary_timeline}
Rui Yan, Xiaojun Wan, Jahna Otterbacher, Liang Kong, Xiaoming Li, and Yan
  Zhang. 2011{\natexlab{b}}.
\newblock \href {https://doi.org/10.1145/2009916.2010016} {Evolutionary
  timeline summarization: A balanced optimization framework via iterative
  substitution}.
\newblock In \emph{Proc.\ of SIGIR}.

\bibitem[{Zellers et~al.(2018)Zellers, Bisk, Schwartz, and
  Choi}]{zellers-etal-2018-swag}
Rowan Zellers, Yonatan Bisk, Roy Schwartz, and Yejin Choi. 2018.
\newblock \href {https://arxiv.org/abs/1808.05326} {{SWAG}: A large-scale
  adversarial dataset for grounded commonsense inference}.
\newblock In \emph{Proc.\ of EMNLP}.

\bibitem[{Zellers et~al.(2019)Zellers, Holtzman, Bisk, Farhadi, and
  Choi}]{zellers-etal-2019-hellaswag}
Rowan Zellers, Ari Holtzman, Yonatan Bisk, Ali Farhadi, and Yejin Choi. 2019.
\newblock \href {https://arxiv.org/abs/1905.07830} {{H}ella{S}wag: Can a
  machine really finish your sentence?}
\newblock In \emph{Proc.\ of ACL}.

\bibitem[{Zhang and Choi(2021)}]{zhang-choi-2021-situatedqa}
Michael~J.Q.\ Zhang and Eunsol Choi. 2021.
\newblock \href {https://arxiv.org/abs/2109.06157} {{S}ituated{QA}:
  Incorporating extra-linguistic contexts into {QA}}.
\newblock In \emph{Proc.\ of EMNLP}.

\bibitem[{Zhang et~al.(2021)Zhang, Ma, Shi, and Lin}]{zhang-etal-2021-mr}
Xinyu Zhang, Xueguang Ma, Peng Shi, and Jimmy Lin. 2021.
\newblock \href {https://aclanthology.org/2021.mrl-1.12} {Mr.\ {T}y{D}i: A
  multi-lingual benchmark for dense retrieval}.
\newblock In \emph{Proc.\ of MRL}.

\end{thebibliography}

\section*{Checklist}


\begin{enumerate}

\item For all authors...
\begin{enumerate}
  \item Do the main claims made in the abstract and introduction accurately reflect the paper's contributions and scope?
    \answerYes{}
  \item Did you describe the limitations of your work?
    \answerYes{}
  \item Did you discuss any potential negative societal impacts of your work?
    \answerYes{}
  \item Have you read the ethics review guidelines and ensured that your paper conforms to them?
    \answerYes{}
\end{enumerate}

\item If you are including theoretical results...
\begin{enumerate}
  \item Did you state the full set of assumptions of all theoretical results?
    \answerNA{}
	\item Did you include complete proofs of all theoretical results?
    \answerNA{}
\end{enumerate}

\item If you ran experiments (e.g. for benchmarks)...
\begin{enumerate}
  \item Did you include the code, data, and instructions needed to reproduce the main experimental results (either in the supplemental material or as a URL)?
    \answerYes{}
  \item Did you specify all the training details (e.g., data splits, hyperparameters, how they were chosen)?
    \answerYes{}
	\item Did you report error bars (e.g., with respect to the random seed after running experiments multiple times)?
    \answerNA{}
	\item Did you include the total amount of compute and the type of resources used (e.g., type of GPUs, internal cluster, or cloud provider)?
    \answerYes{}
\end{enumerate}

\item If you are using existing assets (e.g., code, data, models) or curating/releasing new assets...
\begin{enumerate}
  \item If your work uses existing assets, did you cite the creators?
    \answerYes{}
  \item Did you mention the license of the assets?
    \answerYes{}
  \item Did you include any new assets either in the supplemental material or as a URL?
    \answerYes{}
  \item Did you discuss whether and how consent was obtained from people whose data you're using/curating?
    \answerNA{}
  \item Did you discuss whether the data you are using/curating contains personally identifiable information or offensive content?
    \answerYes{}
\end{enumerate}

\item If you used crowdsourcing or conducted research with human subjects...
\begin{enumerate}
  \item Did you include the full text of instructions given to participants and screenshots, if applicable?
    \answerNA{}
  \item Did you describe any potential participant risks, with links to Institutional Review Board (IRB) approvals, if applicable?
    \answerNA{}
  \item Did you include the estimated hourly wage paid to participants and the total amount spent on participant compensation?
    \answerNA{}
\end{enumerate}

\end{enumerate}


\appendix

\newpage
\begin{appendices}
\section{Baseline Configurations}
\label{appendix:baseline_configs}
We provide the configurations for our real-time baselines (\S\ref{section:baselines}).
We use the open-source, Transformers library and ensure easy replication of our results.
Table \ref{tab:rag-setting} lists the configurations for dense passage retrieval \citep{karpukhin-etal-2020-dense} and retrieval-augmented generation \citep{rag2020}.
We generally follow the default settings from the Transformers library.
Seen in Table \ref{tab:t5-setting} is the configuration for the closed-book T5 baseline. 
We again generally follow the default setting.

\begin{table}[h]
\small
\centering
\caption{Configurations for dense passage retrieval \citep{karpukhin-etal-2020-dense} and retrieval-augmented generation \citep{rag2020} from the Transformers library \citep{wolf-etal-2020-transformers}.}
\vspace{0.5cm}
\begin{tabular}{@{} l@{\hspace{-0.0cm}} r @{}}
\toprule[.1em]
\textbf{Option} & \textbf{Value}\\
\midrule[.1em]
n\_docs & 5\\
max\_combined\_length &  300\\
retrieval\_vector\_size & 768\\
retrieval\_batch\_size & 8\\
is\_encoder\_decoder &  True\\
prefix & None\\
bos\_token\_id &  None\\
pad\_token\_id & None\\
eos\_token\_id & None\\
decoder\_start\_token\_id & None\\
title\_sep & '/' \\
doc\_sep & '//'\\
dataset & 'wiki\_dpr'\\
dataset\_split & 'train'\\
index\_name & 'compressed'\\
index\_path & None\\
passages\_path & None\\
use\_dummy\_dataset & False\\
reduce\_loss & False\\
label\_smoothing & 0.0\\
do\_deduplication & True\\
exclude\_bos\_score & False\\
do\_marginalize & False\\
output\_retrieved & False\\
use\_cache & True\\
forced\_eos\_token\_id & None\\
\bottomrule[.1em]
\end{tabular}
\label{tab:rag-setting}
\end{table}

\begin{table}[h]
\small
\centering
\caption{Configuration for the closed-book T5 baseline \citep{2020t5} from the Transformer library.}
\begin{tabular}{@{} l@{\hspace{-0.5cm}} r @{}}
\toprule[.1em]
\textbf{Option} & \textbf{Value}\\
\midrule[.1em]
\_name\_or\_path & /home/patrick/t5/t5-11b-ssm-nq\\
architectures &  [
    "T5ForConditionalGeneration"]\\
  d\_f &  65536\\
  d\_kv & 128 \\
  d\_model & 1024 \\
  decoder\_start\_token\_id & 0\\
  dropout\_rate & 0.1 \\
  eos\_token\_id & 1\\
  feed\_forward\_proj" & relu \\
  initializer\_factor & 1.0 \\
  is\_encoder\_decoder &  True \\
  layer\_norm\_epsilon &  1e-06\\
  model\_type &  t5\\
  num\_decoder\_layers &  24\\
  num\_heads &  128 \\
  num\_layers &  24 \\
  output\_past &  True \\
  pad\_token\_id & 0\\
  relative\_attention\_num\_buckets &  32\\
  tokenizer\_class & T5Tokenizer\\
  vocab\_size & 32128\\
\bottomrule[.1em]
\end{tabular}
\label{tab:t5-setting}
\end{table}

\section{\realtime Interface}
\label{appendix:interface}
\begin{figure*}[h!]
\centering
    \includegraphics[width=0.90\textwidth]{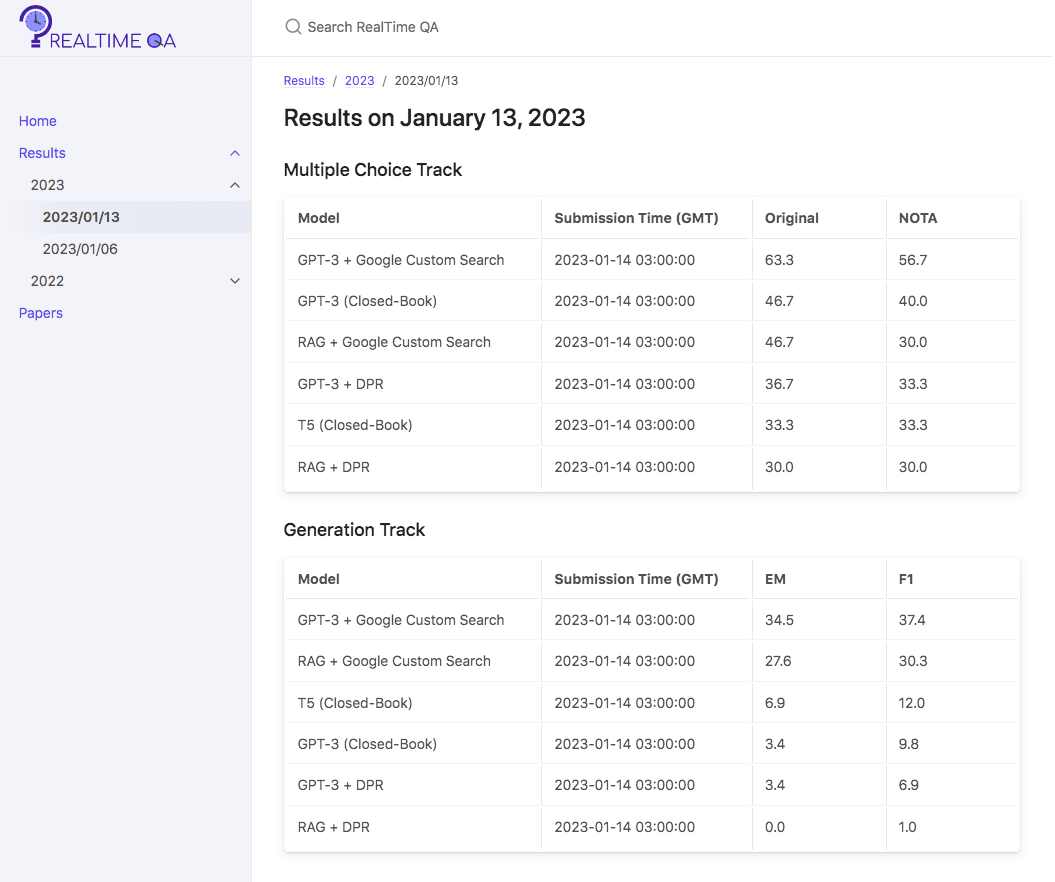}
\caption{\realtime interface. It is updated every week, and all six baselines are evaluated as soon as the questions are available. Submissions will be shown on the same page, together with their submission time.}
\label{fig:interface}
\end{figure*}

Fig.\ \ref{fig:interface} shows our \realtime interface.
It gets updated every week, and all six baselines are evaluated as soon as the questions are available. Submissions will be shown on the same page, together with their submission time. 

\section{\realtime Statistics}
\label{appendix:stats}
Table \ref{tab:stats} provide more detailed statistics of \realtime from the first four weeks.
We analyze the questions from the first four weeks along genres and answer types.
We also found that $\sim$10$\%$ of the questions were not strictly time-sensitive.
These questions include, for example, ``Temperatures in Britain are set to soar this weekend, but what is the hottest UK temperature on record?'' from June 17, 2022. We do not filter out these cases to simulate information-seeking, naturally-occurring scenarios.

\begin{table*}[h]
\centering
\small
\caption{Detailed statistics ($\%$) of \realtime. We analyze the questions from the first four weeks along genres and answer types.
We also found that $\sim$12$\%$ of the questions were not strictly time-sensitive.
These questions include, for example, ``Temperatures in Britain are set to soar this weekend, but what is the hottest UK temperature on record?'' from June 17, 2022. We do not filter out these cases to simulate information-seeking, naturally-occurring scenarios.}
\vspace{0.5cm}
\begin{tabular}{@{}  ccccccc }
\toprule[.1em]

\multicolumn{7}{c}{\textbf{Genre}}
\\
\cmidrule(lr){1-7}
\textbf{Politics}
&\textbf{Business}
& \textbf{Entertain}
&\textbf{Science}
&\textbf{Technology}
&\textbf{Health}
&\textbf{Disaster}
\\

\midrule[.1em]
36.9\%
& 17.5\%
&17.5\%
& 7.0\%
& 7.0\%
& 8.8\%
& 5.2\%

\\
\toprule[.1em]
\multicolumn{7}{c}{\textbf{Answer Type}}
\\
\cmidrule(lr){1-7}
\textbf{Person}
&\textbf{Location}
&\textbf{Time}
&\textbf{Number}
&\textbf{Organization}
&\textbf{Event}
&\textbf{Miscellaneous}

\\
\midrule[.1em]

12.3\%
& 19.2\%
& 5.3\%
& 22.8\%
& 3.5\%
& 8.8\%
& 28.1\%

\\

\bottomrule[.1em]
\end{tabular}
\label{tab:stats}
\end{table*}

\end{appendices}

\end{document}